\documentclass[journal]{IEEEtran}
\ifCLASSINFOpdf
\else
\fi
\usepackage{amsmath}
\usepackage{amssymb}% http://ctan.org/pkg/amssymb
\usepackage{pifont}% http://ctan.org/pkg/pifont

\usepackage{subfig}
\usepackage{comment}

\usepackage{algorithmic}

\usepackage{hyperref}
\hypersetup{
	colorlinks=true,
	linkcolor=red,
	filecolor=magenta,      
	urlcolor=black,
	citecolor=blue
}
\usepackage{booktabs} % For formal tables
\usepackage{mathrsfs}
\usepackage{amsfonts}
\usepackage[ruled]{algorithm2e} % For algorithms

\usepackage{multirow}
\usepackage{graphicx}
\usepackage{color}
\usepackage{balance}
\newtheorem{definition}{Definition}
\begin{document}
\title{ A Comprehensive Survey on Graph Neural Networks}

\author{Zonghan Wu, Shirui Pan,~\IEEEmembership{Member,~IEEE},
Fengwen Chen,
Guodong Long,\\
Chengqi Zhang,~\IEEEmembership{Senior Member,~IEEE}, % <-this % stops a s <-this % 	
Philip S. Yu, ~\IEEEmembership{Fellow,~IEEE}% <-this % stops a s <-this % stops a space

\IEEEcompsocitemizethanks{
	\IEEEcompsocthanksitem Z. Wu,  F. Chen, G. Long, C. Zhang are with Centre for Artificial Intelligence, FEIT, University of Technology Sydney, NSW 2007, Australia (E-mail: zonghan.wu-3@student.uts.edu.au; fengwen.chen@student.uts.edu.au;   guodong.long@uts.edu.au; chengqi.zhang@uts.edu.au).
	\IEEEcompsocthanksitem S. Pan is with Faculty of Information Technology, Monash University, Clayton, VIC 3800, Australia (Email: shirui.pan@monash.edu). 
	\IEEEcompsocthanksitem P. S. Yu is with Department of Computer Science, University of Illinois at Chicago, Chicago, IL 60607-7053, USA (Email: psyu@uic.edu)
	\IEEEcompsocthanksitem Corresponding author: Shirui Pan.
	%\IEEEcompsocthanksitem L. Yao is with the School of Computer Science and Engineering, The University of New South Wales, Sydney, NSW 2052, Australia (e-mail: lina.yao@unsw.edu.au).
}

\thanks{Manuscript received Dec xx, 2018; revised Dec xx, 201x.}}


\markboth{Journal of \LaTeX\ Class Files,~Vol.~xx, No.~xx, August~2019}%
{Shell \MakeLowercase{\textit{et al.}}: Bare Demo of IEEEtran.cls for IEEE Journals}



\maketitle

\begin{abstract}
Deep learning has revolutionized many machine learning tasks in recent years, ranging from image classification and video processing to speech recognition and natural language understanding. The data in these tasks are typically represented in the Euclidean space. However, there is an increasing number of applications where data are generated from non-Euclidean do- mains and are represented as graphs with complex relationships and interdependency between objects. The complexity of graph data has imposed significant challenges on existing machine learning algorithms. Recently, many studies on extending deep learning approaches for graph data have emerged. In this survey, we provide a comprehensive overview of graph neural networks (GNNs) in data mining and machine learning fields.  We propose a new taxonomy to divide the state-of-the-art graph neural networks into four categories, namely recurrent graph neural networks, convolutional graph neural networks, graph autoencoders, and spatial-temporal graph neural networks. We further discuss the applications of graph neural networks across various domains and summarize the open source codes, benchmark data sets, and model evaluation of graph neural networks. Finally, we propose potential research directions in this rapidly growing field.
\end{abstract}


% \fi
\IEEEpeerreviewmaketitle

	%\IEEEraisesectionheading{
\section{Introduction}\label{sec:introduction}


\IEEEPARstart{T}{he} recent success of neural networks has boosted research on pattern recognition and data mining.  Many machine learning tasks such as object detection \cite{redmon2016you,ren2015faster}, machine translation \cite{luong2015effective,wu2016google}, and speech recognition \cite{hinton2012deep}, which once heavily relied on handcrafted feature engineering to extract informative feature sets, has recently been revolutionized by various end-to-end deep learning paradigms, e.g.,  convolutional neural networks (CNNs) \cite{lecun1995convolutional}, recurrent neural networks (RNNs) \cite{hochreiter1997long}, and autoencoders \cite{vincent2010stacked}. The success of deep learning in many domains is partially attributed to the rapidly developing computational resources (e.g., GPU), the availability of big training data, and the effectiveness of deep learning to extract latent representations from Euclidean data (e.g., images, text, and videos). Taking image data as an example,  we can represent an image as a regular grid in the Euclidean space.  A convolutional neural network (CNN) is able to exploit the shift-invariance, local connectivity, and compositionality of image data \cite{bronstein2017geometric}. As a result, CNNs can extract local meaningful features that are shared with the entire data sets for various image analysis.  
    
While deep learning effectively captures hidden patterns of Euclidean data, there is an increasing number of applications where data are 
represented in the form of graphs. 
%Graphs describe the connections, relations, or interactions of objects in a system. For examples, a molecule graph represents a molecule with atoms as nodes and chemical bonds as edges. A citation network describes the citation relationship between academic papers. A recommendation system records the interactions between users and products. The complexity of graph data has imposed significant challenges on existing machine learning algorithms. 
For examples, in e-commence, a graph-based learning system can exploit the interactions between users and products to make highly accurate recommendations. In chemistry, molecules are modeled as graphs, and their bioactivity needs to be identified for drug discovery. In a citation network, papers are linked to each other via citationships and they need to be categorized into different groups. The complexity of graph data has imposed significant challenges on existing machine learning algorithms. 
As graphs can be irregular, a  graph  may  have  a  variable size of unordered nodes, and nodes from a graph may have a different number of neighbors, resulting in some important operations (e.g., convolutions) being  easy to  compute in the image domain, but difficult to apply to the graph domain.  Furthermore, a core assumption of existing machine learning algorithms is that instances are independent of each other. This assumption no longer holds for graph data because each instance (node) is related to others by links of various types, such as  citations, friendships, and interactions.

\begin{figure}[]
\centering
\subfloat[2D Convolution. Analogous to a graph, each pixel in an image is taken as a node where neighbors are determined by the filter size. The 2D convolution takes the weighted average of pixel values of the red node along with its neighbors. The neighbors of a node are ordered and have a fixed size. ]{\includegraphics[width=1.5in]{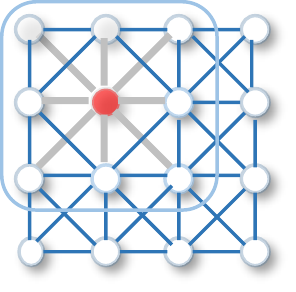}%
\label{fig_first_case}}
\hfill
\subfloat[Graph Convolution. To get a hidden representation of the red node, one simple solution  of the graph convolutional operation is to take the average value of the node features of the red node along with its neighbors. Different from image data, the neighbors of a node are unordered and variable in size.]{\includegraphics[width=1.5in]{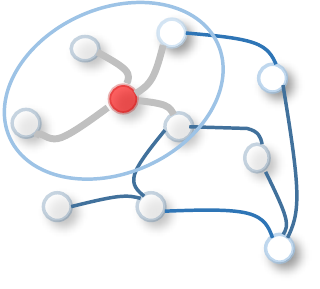}%
\label{fig_second_case}}
\caption{2D Convolution vs. Graph Convolution.}
\label{gcn_cnn}
\end{figure}

Recently, there is increasing interest in extending deep learning approaches for graph data. Motivated by CNNs, RNNs, and autoencoders from deep learning, new general- izations and definitions of important operations have been rapidly developed over the past few years to handle the com- plexity of graph data. For example, a graph convolution can be generalized from a 2D convolution. As illustrated in Figure \ref{gcn_cnn}, an image can be considered as a special case of graphs where pixels are connected by adjacent pixels. Similar to 2D convolution, one may perform graph convolutions by  taking  the  weighted  average  of  a  node's  neighborhood
information.

There are a limited number of existing reviews on the topic of graph neural networks (GNNs). Using the term \textit{geometric deep learning}, Bronstein et al. \cite{bronstein2017geometric} give an overview of deep learning methods in the non-Euclidean domain, including graphs and manifolds. Although it is the first review on GNNs, this survey mainly  reviews  convolutional  GNNs.  Hamilton et al. \cite{hamilton2017representation} cover a limited number of GNNs with a focus on addressing the problem of network embedding.
Battaglia et al. \cite{battaglia2018relational} position \textit{graph networks} as the building blocks for learning from relational data, reviewing part of GNNs under a unified framework. 
%However, their generalized framework is highly abstract, losing insights on each method from its original paper. 
Lee et al. \cite{lee2018attention} conduct a partial survey of GNNs which apply different attention mechanisms.
%Most recently, Zhang et al. \cite{zhang2018deep} present a most up-to-date survey on deep learning for graphs, missing those studies on graph generative and spatial-temporal networks.
In summary, existing surveys only include some of the GNNs and examine a limited number of works, thereby missing the most recent development of GNNs. Our survey provides a comprehensive overview of GNNs, for both interested researchers who want to enter this rapidly developing field and experts who would like to compare GNN models.
To cover a broader range of methods, this survey considers GNNs as all deep learning approaches for graph data.

\vspace{.2cm}
\textbf{Our contributions}
Our paper makes notable contributions summarized as follows:
\begin{itemize}
\item \textbf{New taxonomy} We propose a new taxonomy of graph neural networks.  Graph neural networks are categorized into four groups: recurrent graph neural networks, convolutional graph neural networks, graph autoencoders, and spatial-temporal graph neural networks. 

\item \textbf{Comprehensive review} We provide the most comprehensive overview of modern deep learning techniques for graph data. For each type of graph neural network, we provide detailed descriptions on representative models, make the necessary comparison, and summarise the corresponding algorithms.

\item \textbf{Abundant resources} We collect abundant resources on graph neural networks,  including state-of-the-art models, benchmark data sets, open-source codes, and practical applications. This survey can be used as a hands-on guide for understanding, using, and developing different deep learning approaches for various real-life applications. 

\item \textbf{Future directions} We discuss theoretical aspects of graph neural networks, analyze the limitations of existing methods, and suggest four possible future research directions in terms of model depth, scalability trade-off, heterogeneity, and dynamicity.
\end{itemize}
	
\vspace{.2cm}
\textbf{Organization of our survey}
The rest of this survey is organized as follows. Section \ref{sec:definition} outlines the background of graph neural networks, lists commonly used notations, and defines graph-related concepts. Section \ref{sec:categorization} clarifies the categorization of graph neural networks. Section \ref{sec:grn}-\ref{sec:stgcn} provides an overview of graph neural network models. Section \ref{sec:applications} presents a collection of applications across various domains.  Section \ref{sec:fucture} discusses the current challenges and suggests future directions. Section \ref{sec:conclusion} summarizes the paper.

\section{Background \& Definition}\label{sec:definition}
In this section, we outline the background of graph neural networks, list commonly used notations, and define graph-related concepts.

\subsection{Background}

\vspace{2mm}
\textbf{A brief history of graph neural networks (GNNs)} Sperduti et al. (1997) \cite{sperduti1997supervised} first applied neural networks to directed acyclic graphs, which motivated early studies on GNNs. The notion of graph neural networks was initially outlined in Gori et al. (2005) \cite{gori2005new} and further elaborated in Scarselli et al. (2009) \cite{scarselli2009graph}, and Gallicchio et al. (2010) \cite{gallicchio2010graph}. 
These early studies fall into the category of recurrent graph neural networks (RecGNNs). They learn a target node's representation by propagating neighbor information in an iterative manner until a stable fixed point is reached. 
This process is computationally expensive, and recently there have been increasing efforts to overcome these challenges \cite{li2015gated,dai2018learning}. 

Encouraged by the success of CNNs in the computer vision domain, a large number of methods that re-define the notion of \textit{convolution} for graph data are developed in parallel. These approaches are under the umbrella of convolutional graph neural networks (ConvGNNs). ConvGNNs are divided into two main streams, the spectral-based approaches and the spatial-based approaches.  The first prominent research on spectral-based ConvGNNs was presented by Bruna et al. (2013) \cite{bruna2013spectral}, which developed a graph convolution based on the spectral graph theory.  Since this time, there have been increasing improvements, extensions, and approximations on spectral-based ConvGNNs \cite{henaff2015deep, defferrard2016convolutional, kipf2017semi, levie2017cayleynets}.  The research of spatial-based ConvGNNs started much earlier than spectral-based ConvGNNs. In 2009, Micheli et al. \cite{micheli2009neural} first addressed graph mutual dependency by architecturally composite non-recursive layers while inheriting ideas of message passing from RecGNNs.  However,  the importance of this work was overlooked. Until recently, many spatial-based ConvGNNs (e.g., \cite{atwood2016diffusion,niepert2016learning,gilmer2017neural}) emerged. %due to the fact that spectral methods usually handle the whole graph simultaneously and are difficult to parallelise or scale to large graphs. 
%These methods directly perform graph convolutions by aggregating a node's neighboring information. 
%Together with sampling strategies,  the computation can be operated in a batch of nodes instead of the whole graph \cite{hamilton2017inductive,gao2018large}, which improves the training efficiency.  
The timeline of representative RecGNNs and ConvGNNs is shown in the first column of Table \ref{tab:pub}.
Apart from RecGNNs and ConvGNNs, many alternative GNNs have been developed in the past few years, including graph autoencoders (GAEs) and  spatial-temporal graph neural networks (STGNNs). These learning frameworks can be built on RecGNNs, ConvGNNs, or other neural architectures for graph modeling. 
Details on the categorization of these methods are given in Section \ref{sec:categorization}. 
	
\vspace{.1cm}
\textbf{Graph neural networks vs. network embedding } 
The research on GNNs is closely related to graph embedding or network embedding, another topic which attracts increasing attention from both the data mining and machine learning communities \cite{hamilton2017representation,cui2017survey,zhang2018network,cai2018comprehensive,goyal2018graph,pan2016tri}.  Network embedding aims at representing network nodes as low-dimensional vector representations, preserving both network topology structure and node content information, so that any subsequent graph analytics task such as classification, clustering, and recommendation  can be easily performed using simple off-the-shelf machine learning algorithms (e.g., support vector machines for classification). 
Meanwhile, GNNs are deep learning models aiming at addressing graph-related tasks in an end-to-end manner. Many GNNs explicitly extract high-level representations. %However, node representations obtained by graph neural networks are specific to a certain task while node representations obtained by network embedding are general to all subsequent tasks.
The main distinction between GNNs and network embedding is that GNNs are a group of neural network models which are designed for various tasks while network embedding covers various kinds of methods targeting the same task. Therefore, GNNs can address the network embedding problem through a graph autoencoder framework. On the other hand, network embedding contains other non-deep learning methods such as  matrix factorization \cite{IJCAI-18-Shen,ICDM-18-Hong} and random walks \cite{perozzi2014deepwalk}.
	
\vspace{.1cm}
\textbf{Graph neural networks vs. graph kernel methods}
Graph kernels are historically dominant techniques to solve the problem of graph classification \cite{vishwanathan2010graph,shervashidze2011weisfeiler,navarin2017approximated}. These methods employ a kernel function to measure the similarity between pairs of graphs so that kernel-based algorithms like support vector machines can be used for supervised learning on graphs.  Similar to GNNs, graph kernels can embed graphs or nodes into vector spaces by a mapping function. The difference is that this mapping function is deterministic rather than learnable. Due to a pair-wise similarity calculation, graph kernel methods suffer significantly from computational bottlenecks. GNNs, on one hand, directly perform graph classification based on the extracted graph representations and therefore are much more efficient than graph kernel methods. For a further review of graph kernel methods, we refer the readers to \cite{kriege2019survey}.%\cite{gartner2003survey,kriege2019survey}.

	\begin{table}[t]
		\caption{Commonly used notations.}
		\label{tab:notations}
		\centering
		\begin{tabular} {  l l p{7cm} } \toprule
				\textbf{Notations}& \textbf{Descriptions} \\ \midrule
			    $|\cdot|$ & The length of a set. \\ \hline
				$\odot$ & Element-wise product. \\ \hline
				$G$& A graph. \\ \hline
				$V$& The set of nodes in a graph.\\ \hline
				$v$ & A node $v\in V$. \\ \hline
			    $E$& The set of edges in a graph.\\ \hline
				$e_{ij}$ & An edge $e_{ij}\in E$.\\ \hline
				$N(v)$ & The neighbors of a node $v$. \\ \hline
				$\mathbf{A}$ & The graph adjacency matrix.  \\ \hline
				$\mathbf{A}^T$ & The transpose of the matrix
				$\mathbf{A}$. \\ \hline
				$\mathbf{A}^n, n\in Z$ & The $n^{th}$ power of
				$\mathbf{A}$. \\ \hline
				$[\mathbf{A},\mathbf{B}]$ & The concatenation of $\mathbf{A}$ and $\mathbf{B}$. \\ \hline
                $\mathbf{D}$ & The degree matrix of $\mathbf{A}$. $\mathbf{D}_{ii} = \sum_{j=1}^n \mathbf{A}_{ij}$. \\ \hline
				$n$ & The number of nodes, $n = |V|$. \\ \hline
				$m$ & The number of edges, $m = |E|$. \\ \hline
				$d$  & The dimension of a node feature vector.\\ \hline
				$b$ & The dimension of a hidden node feature vector. \\ \hline
			    $c$  & The dimension of an edge feature vector.\\ \hline
				$\mathbf{X} \in \mathbf{R}^{n\times d}$ & The feature matrix of a graph. \\ \hline
				$\mathbf{x} \in \mathbf{R}^n$ & The feature vector of a graph in the case of $d=1$. \\ \hline
				$\mathbf{x}_v \in \mathbf{R}^d$ & The feature vector of the node $v$. \\ \hline
				$\mathbf{X}^e \in \mathbf{R}^{m\times c}$ & The edge feature matrix of a graph. \\ \hline
				$\mathbf{x}^e_{(v,u)} \in \mathbf{R}^{c}$ & The edge feature vector of the edge $(v,u)$. \\ \hline
				$\mathbf{X}^{(t)} \in \mathbf{R}^{n\times d}$ & The node feature matrix of a graph at the time step $t$. \\ \hline
				$\mathbf{H} \in \mathbf{R}^{n \times b}$ & The node hidden feature matrix. \\ \hline
				$\mathbf{h}_v \in \mathbf{R}^{b}$ & The hidden feature vector of node $v$. \\ \hline
				$k$ & The layer index \\ \hline
				$t$ & The time step/iteration index  \\ \hline
				$\sigma(\cdot)$ & The sigmoid activation function. \\ \hline
				$\sigma_h(\cdot)$ & The tangent hyperbolic activation function. \\ \hline	$\mathbf{W},\mathbf{\Theta},w,\theta$ & Learnable model parameters. \\  
				\bottomrule
		\end{tabular}
	\end{table}

	\subsection{Definition}

	Throughout this paper, we use bold uppercase characters to denote matrices and  bold lowercase characters denote vectors. Unless particularly specified, the notations used in this paper are illustrated in Table \ref{tab:notations}. Now we define the minimal set of definitions required to understand this paper. 
	
	\begin{definition}[Graph]
		A graph is represented as $G$ = $(V,E)$ where $V$ is the set of vertices or nodes (we will use nodes throughout the paper), and $E$ is the set of edges. Let $v_i \in V$ to denote a node and $e_{ij}=(v_i,v_j) \in E$ to denote an edge pointing from $v_j$ to $v_i$. The neighborhood of a node $v$ is defined as $N(v)= \{u\in V|(v,u)\in E\}$.
		The adjacency matrix $\mathbf{A}$ is a $n\times n$ matrix with $A_{ij} = 1$ if $e_{ij} \in E$ and $A_{ij} = 0$ if $e_{ij} \notin E$. A graph may have node attributes $\mathbf{X}$ \footnote{Such graph is referred to an \textit{attributed graph} in literature.}, where $\mathbf{X} \in \mathbf{R}^{n\times d}$ is a node feature matrix with $\mathbf{x}_v \in \mathbf{R}^d$ representing the feature vector of a node $v$.   Meanwhile, a graph may have edge attributes $\mathbf{X}^e$, where $\mathbf{X}^e \in \mathbf{R}^{m\times c}$ is an edge feature matrix with $\mathbf{x}^e_{v,u} \in \mathbf{R}^c$ representing the feature vector of an edge $(v,u)$.
	
	\end{definition}

%In the case of $d=1$, we use $\mathbf{x} \in \mathbf{R}^n$ to denote the feature vector of the graph.  When edge features are available, we use $\mathbf{X^e} \in \mathbf{R}^{m\times c}$ to denote edge features.
%The degree of a node $D(v_i)=\sum_j \mathbf{A}_{ij}$. 
	
	\begin{definition}[Directed Graph]
		A directed graph is a graph with all edges directed from one node to another. An undirected graph is considered as a special case of directed graphs where there is a pair of edges with inverse directions if two nodes are connected. A graph is undirected if and only if the adjacency matrix is symmetric.
		%The adjacency matrix of a directed graph can be asymmetric, i.e., $\mathbf{A}\ne \mathbf{A}^T$. The adjacency matrix of a undirected graph is symmetric, i.e., $\mathbf{A}=\mathbf{A}^T$. 
	\end{definition}
	
	%\begin{definition}[Weighted Graph]
	%A weighted graph is a graph where each edge of the graph is associated with a weight. For a weighted graph, $A_{ij}=w_{ij}>0$ if $e_{ij} \in E$.
	%\end{definition}
	
	%\begin{definition}[Heterogeneous Graph]
	%A heterogeneous graph is a graph with multiple type of nodes and/or mutliple type of edges. Formally it is defined as $\mathcal{G}$ = $(V,E,A,Z_v,Z_e)$  where $Z_v$ maps a node to its node type and $Z_e$ maps an edge to its edge type.  Where there only exists one node type and one edge type, the graph is homogeneous otherwise it is heterogeneous. 
	%\end{definition}
	
	\begin{definition}[Spatial-Temporal Graph]
		A spatial-temporal graph is an attributed graph where the node attributes change dynamically over time.  The spatial-temporal graph is defined as $G^{(t)}$ = $(\mathbf{V},\mathbf{E},\mathbf{X}^{(t)})$ with $\mathbf{X}^{(t)} \in \mathbf{R}^{n\times d}$.
	\end{definition}

	\section{Categorization and  Frameworks}\label{sec:categorization}
\begin{figure}[htp]
		\centering
% 		\subfloat[The Process of Graph Convolution Networks for Node Representation \cite{kipf2017semi}.  A GCN layer encapsulates each node's hidden representation by aggregating feature information from its neighbors. After feature aggregation, a non-linear transformation is applied to the resultant outputs. By stacking multiple layers, the final hidden representation of each node receives messages from further neighborhood. ]{\includegraphics[width=3.5in]{fig/gcn.png}%
% 			\label{fig:gcn}}
% 		\hfill

        \subfloat[A ConvGNN with multiple graph convolutional layers.  A graph convolutional layer encapsulates each node's hidden representation by aggregating feature information from its neighbors. After feature aggregation, a non-linear transformation is applied to the resulted outputs. By stacking multiple layers, the final hidden representation of each node receives messages from a further neighborhood.]{\includegraphics[width=3.5in]{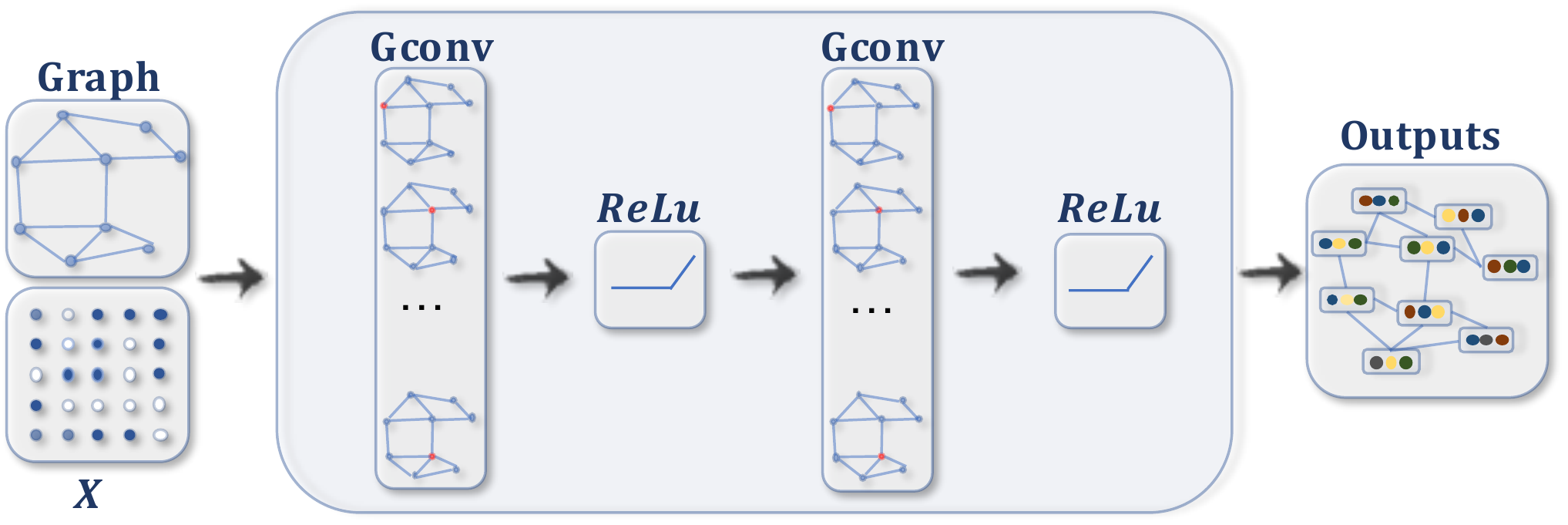}%
			    \label{fig:gcn}
        }
        \hfill
		\subfloat[A ConvGNN with pooling and readout layers for graph classification \cite{defferrard2016convolutional}. A graph convolutional layer is followed by a pooling layer to coarsen a graph into sub-graphs so that node representations on coarsened graphs represent higher graph-level representations. A readout layer summarizes the final graph representation by taking the sum/mean of hidden representations of sub-graphs. ]{\includegraphics[width=3.5in]{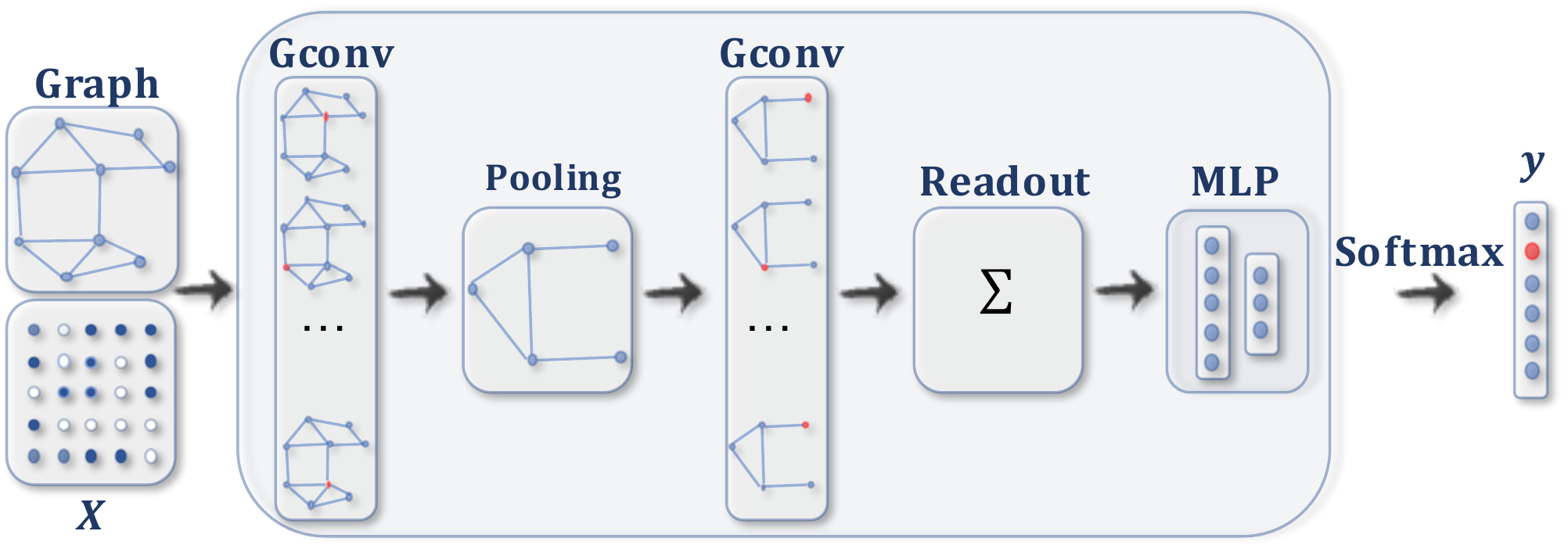}%
			\label{fig:gcnpool}}
		\hfill
		\subfloat[A GAE for network embedding \cite{kipf2016variational}. The encoder uses graph convolutional layers to get a network embedding for each node. The decoder computes the pair-wise distance given network embeddings. After applying a non-linear activation function, the decoder reconstructs the graph adjacency matrix. The network is trained by minimizing the discrepancy between the real adjacency matrix and the reconstructed adjacency matrix.   ]{\includegraphics[width=3.5in]{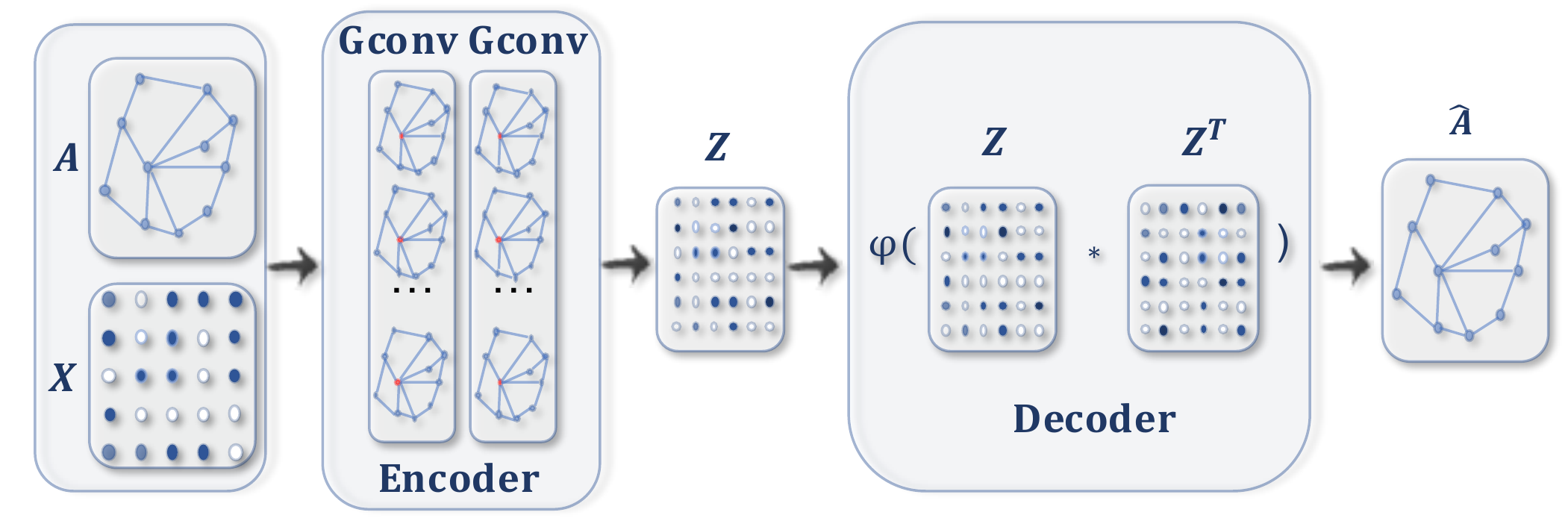}%
			\label{fig:gae}}
		%\hfill
		%\subfloat[Graph Generative Networks with GCN \cite{de2018molgan}. A generator first samples an initial vector from a standard normal distribution. Passing this initial vector through a neural network, the generator outputs a dense adjacency matrix $A$ and a corresponding feature matrix $X$.  Next, the generator produces a sampled discrete $\tilde{A}$ and $\tilde{X}$ from categorical distributions based on $A$ and $X$. Finally, GCN is used to derive a vector representation of the sampled graph. Feeding this graph representation to two distinct neural networks, a discriminator and a reward network outputs a score between zero and one separately, which will be used as feedback to update the model parameters.. ]{\includegraphics[width=3.5in]{fig/molgan.png}%
			%\label{fig:ggn}}
		\hfill
		\subfloat[A STGNN for spatial-temporal graph forecasting \cite{yuspatio}. A graph convolutional layer is followed by a 1D-CNN layer. The graph convolutional layer operates on $A$ and $X^{(t)}$ to capture the spatial dependency, while the 1D-CNN layer slides over $X$ along the time axis to capture the temporal dependency. The output layer is a linear transformation, generating a prediction for each node, such as its future value at the next time step. ]{\includegraphics[width=3.5in]{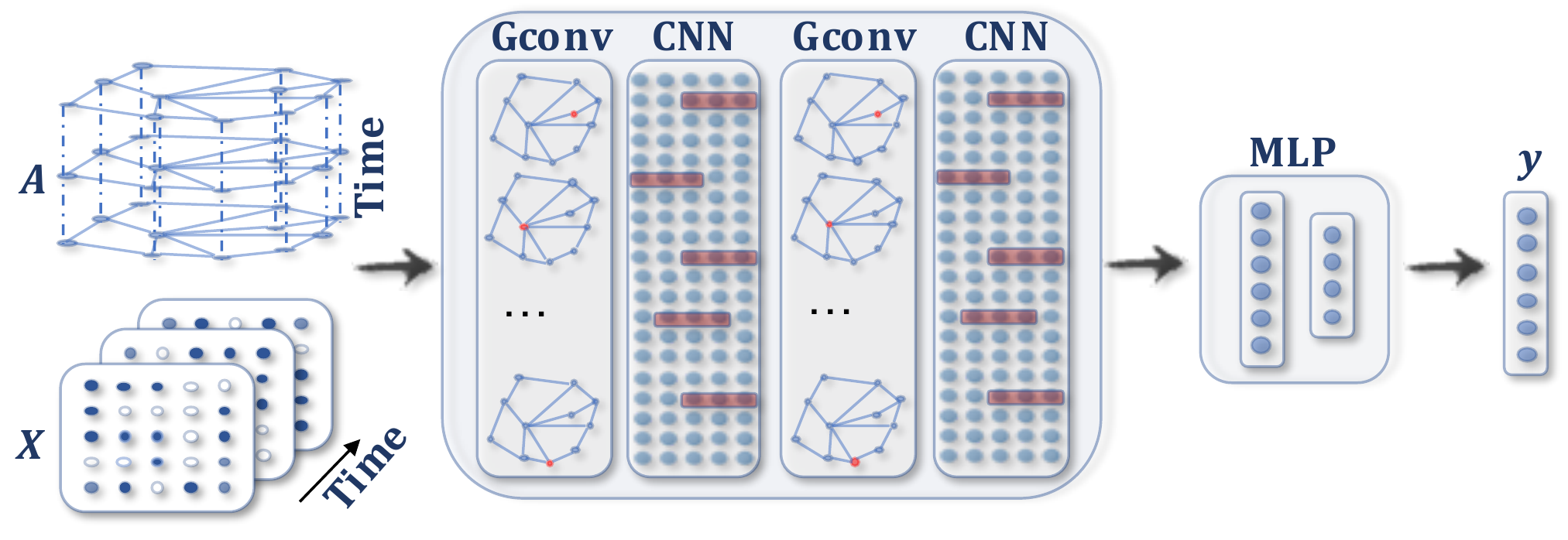}%
			\label{fig:gst}}
		\caption{Different graph neural network models built with graph convolutional layers. The term Gconv denotes a graph convolutional layer. The term MLP denotes a multi-layer perceptron. The term CNN denotes a standard convolutional layer. }
		\label{manygcn}
	\end{figure}

	In this section, we present our taxonomy of graph neural networks (GNNs), as shown in Table \ref{tab:pub}. We categorize graph neural networks (GNNs) into recurrent graph neural  networks (RecGNNs), convolutional graph neural networks (ConvGNNs), graph autoencoders (GAEs), and  spatial-temporal graph neural networks (STGNNs). Figure \ref{manygcn} gives examples of various model architectures. %Among these categories we put emphasis on convolutional graph neural networks as many other models under graph autoencoders and spatial-temporal graph neural networks can utilize graph convolutional layers as building blocks, as demonstrated in Figure \ref{manygcn}. 
	In the following, we give a brief introduction of each category.
	%play a central role in capturing structural dependencies. 

	\begin{table*}[htb]
		\caption{Taxonomy and representative publications of Graph Neural Networks (GNNs)}
		\label{tab:pub}
		\centering
		\begin{tabular}{l l l}
			\toprule
			\multicolumn{2}{l}{Category}                               & Publications \\ \midrule
			\multicolumn{2}{l}{Recurrent Graph Neural Networks (RecGNNs)} & \cite{scarselli2009graph, gallicchio2010graph,li2015gated,dai2018learning} \\ \midrule
			\multirow{2}{*}{} & Spectral methods &      \cite{bruna2013spectral,henaff2015deep,defferrard2016convolutional,kipf2017semi,levie2017cayleynets,li2018adaptive,zhuang2018dual}        \\ \cline{2-3} 
			Convolutional Graph Neural Networks (ConvGNNs) & Spatial methods  & \begin{tabular}[c]{@{}l@{}} \cite{micheli2009neural,atwood2016diffusion,niepert2016learning,gilmer2017neural,hamilton2017inductive,velickovic2017graph,monti2017geometric}\\
			\cite{gao2018large,tran2018filter,bacciu2018contextual,zhang2018gaan,chen2018fastgcn,chen2018stochastic,huang2018ada}   \\
			\cite{zhang2018end,li2018deeper,ying2018hierarchical,liu2018geniepath,velivckovic2019deep,xu2019how,chiang2019cluster}\\
			\end{tabular}\\ \midrule
			\multirow{2}{*}{Graph Autoencoders (GAEs)} & Network Embedding &  
             \cite{cao2016deep,wang2016sdne,kipf2016variational,pan2018adversarially,tu2018deep,yu2018learning}          \\ \cline{2-3} 
             & Graph Generation &  \cite{    li2018learning,you2018graphrnn,simonovsky2018graphvae,ma2018constrained,de2018molgan,bojchevski2018netgan}\\ \midrule
			\multicolumn{2}{l}{Spatial-temporal Graph Neural Networks (STGNNs)}        &     \cite{seo2018structured,li2018diffusion,jain2016structural,yuspatio,yan2018spatial,wu2019graph,guo2019att}         \\ \bottomrule
		\end{tabular}
	\end{table*}

    \subsection{Taxonomy of Graph Neural Networks (GNNs)}
    %\begin{itemize}
    %\vspace{2mm}
    \noindent\textbf{Recurrent graph neural networks (RecGNNs)} 
    mostly are pioneer works of graph neural networks. RecGNNs aim to learn node representations with recurrent neural architectures. They assume a node in a graph constantly exchanges information/message with its neighbors until a stable equilibrium is reached. %While conceptually important, RecGNNs lead later research on convolutional graph neural networks. 
    RecGNNs are conceptually important and inspired later research on convolutional graph neural networks.
    In particular, the idea of message passing is inherited by spatial-based convolutional graph neural networks. 

    \vspace{1mm}
    \noindent\textbf{Convolutional graph neural networks (ConvGNNs)} generalize the operation of \textit{convolution} from grid data to graph data. The main idea is to generate a node $v$'s representation by aggregating its own features  $\mathbf{x}_v$ and neighbors' features $\mathbf{x}_u$, where $u \in N(v)$.
    Different from RecGNNs, ConvGNNs stack multiple graph convolutional layers to extract  high-level node representations. 
    %The key is to learn a function $f$ to generate a node $v$'s representation by aggregating its own features  $\mathbf{x}_v$ and neighbors' features $\mathbf{x}_u$, where $u \in N(v)$.  
    ConvGNNs play a central role in building up many other complex GNN models. Figure \ref{fig:gcn} shows a ConvGNN for node classification. Figure \ref{fig:gcnpool} demonstrates a ConvGNN for graph classification.

    \vspace{1mm}\noindent\textbf{Graph autoencoders (GAEs)} are unsupervised learning frameworks which encode nodes/graphs into a latent vector space and reconstruct graph data from the encoded information. GAEs are used to learn network embeddings and graph generative distributions. For network embedding, GAEs learn latent node representations through reconstructing graph structural information such as the graph adjacency matrix. For graph generation, some methods generate nodes and edges of a graph step by step while other methods output a graph all at once. Figure \ref{fig:gae} presents a GAE for network embedding.

    %For plain graphs, many algorithms directly prepossess the adjacency matrix, by either constructing a new matrix (i.e., pointwise mutual information matrix) with rich information \cite{cao2016deep} or feeding the adjacency matrix to an autoencoder model and capturing both first order and second order information \cite{wang2016sdne}.   For attributed graphs,   GAE models tend to employ GCN \cite{kipf2017semi} as a building block for the encoder and reconstruct the structure information via a link prediction decoder \cite{kipf2016variational,pan2018adversarially}.
	%\vspace{2mm}\noindent\textbf{Graph Generative Networks} aim to generate  plausible structures from data. Generating graphs given a graph empirical distribution is fundamentally challenging, mainly because graphs are complex data structures. To address this problem, researchers have explored to factor the generation process as forming nodes and edges alternatively \cite{you2018graphrnn,li2018learning}, to employ generative adversarial training \cite{de2018molgan,bojchevski2018netgan}. One promising application domain of graph generative networks is chemical compound synthesis. In a chemical graph, atoms are treated as nodes and chemical bonds are treated as edges. The task is to discover new synthesizable molecules which possess certain chemical and physical properties.
	
    \vspace{1mm}\noindent\textbf{Spatial-temporal graph neural networks (STGNNs)} aim to learn hidden patterns from spatial-temporal graphs, which become increasingly important in a variety of applications such as traffic speed forecasting \cite{li2018diffusion}, driver maneuver anticipation \cite{jain2016structural}, and  human action recognition \cite{yan2018spatial}. 
    %For example, the traffic network is a natural graph with each key location as a node whose traffic data is continuously monitored. By developing effective graph spatial-temporal network models, one can accurately predict the traffic status over the whole traffic system \cite{li2018diffusion,yuspatio}. 
    The key idea of STGNNs is to consider spatial dependency and temporal dependency at the same time. Many current approaches integrate graph convolutions to capture spatial dependency with RNNs or CNNs to model the temporal dependency. Figure \ref{fig:gst} illustrates a STGNN for spatial-temporal graph forecasting.
	%\end{itemize}

	\subsection{Frameworks}
%    As ConvGNNs play a central role in building up modern graph neural networks, we summarize frameworks of ConvGNNs in the following. 
With the graph structure and node content information as inputs,  the outputs of  GNNs can focus on different graph analytics tasks with one of the following mechanisms:
	\begin{itemize}
		\item \textbf{Node-level} outputs relate to node regression and node classification tasks. RecGNNs and ConvGNNs can extract high-level node representations by information propagation/graph convolution. With a multi-perceptron or a softmax layer as the output layer, GNNs are able to perform node-level tasks in an end-to-end manner.
		%\item \textbf{Node-level} outputs relate to node regression and classification tasks. This is achieved by stacking multiple graph convolutional layers with non-linear activation functions, followed by a multi-perceptron or softmax layer as the output layer. We review graph convolutional modules in Section \ref{sec:spectral_gcn} and Section \ref{sec:spatial_gcn}.
		
		\item \textbf{Edge-level} outputs relate to the edge classification and link prediction tasks.  With two nodes' hidden representations from GNNs as inputs, a similarity function or a neural network can be utilized to predict the label/connection strength of an edge.
		
		%\item \textbf{Graph-level} outputs relate to the graph classification task. To obtain a compact representation on graph level, a pooling module is used to coarse a graph into sub-graphs or to sum/average over the node representations. We review the graph pooling module in Section \ref{sec:pooling}.
		
		% Victor
		\item \textbf{Graph-level} outputs relate to the graph classification task. To obtain a compact representation on the graph level, GNNs are often combined with pooling and readout operations. Detailed information about pooling and readouts will be reviewed in Section \ref{sec:pooling}.
	\end{itemize}
    
    \vspace{2mm}
    \textit{Training Frameworks.} Many GNNs (e.g., ConvGNNs) can be trained in a (semi-) supervised or purely unsupervised way within an end-to-end learning framework, depending on the learning tasks and label information available at hand.
    \begin{itemize}
        \item \textbf{Semi-supervised learning for node-level classification.} Given a single network with partial nodes being labeled and others remaining unlabeled, ConvGNNs can learn a robust model that effectively identifies the class labels for the unlabeled nodes \cite{kipf2017semi}. To this end, an end-to-end framework can be built by stacking a couple of graph convolutional layers followed by a softmax layer for multi-class classification. 
        
        %\item \textbf{Supervised learning for graph-level classification.} Given a graph dataset, graph-level classification aims to predict the class label(s) for an entire graph \cite{zhang2018end,ying2018hierarchical,pan2016joint,pan2017task}. The end-to-end learning for this task can be done with a framework which combines both graph convolutional layers and the pooling procedure \cite{zhang2018end, ying2018hierarchical}. Specifically, by applying graph convolutional layers, we obtain representation with a fixed number of dimensions for each node in every single graph. Then, we can get the representation of an entire graph through pooling which summarizes the representation vectors of all nodes in a graph. Finally, by applying linear layers and a softmax layer, we can build an end-to-end framework for graph classification. An example is given in Fig \ref{fig:gcnpool}.
        
        % Victor
        \item \textbf{Supervised learning for graph-level classification.} Graph-level classification aims to predict the class label(s) for an entire graph \cite{zhang2018end,ying2018hierarchical,pan2016joint,pan2017task}. The end-to-end learning for this task can be realized with a combination of graph convolutional layers, graph pooling layers, and/or readout layers. While graph convolutional layers are responsible for exacting high-level node representations, graph pooling layers play the role of down-sampling, which coarsens each graph into a sub-structure each time. A readout layer collapses node representations of each graph into a graph representation. 
        %Specifically, by applying graph convolutional layers, we obtain a representation with a fixed number of dimensions for each node in each single graph. Then, placing pooling layer in-between graph convolutional layers coarsens each graph into a sub-graph each time. Finally, we can get the representation of an entire graph through a readout operation which summarizes the representation vectors of all nodes in a coarsened sub-graph.
        By applying a multi-layer perceptron and a softmax layer to graph representations, we can build an end-to-end framework for graph classification. An example is given in Fig \ref{fig:gcnpool}.
        
        \item \textbf{Unsupervised learning for graph embedding.} When no class labels are available in graphs, we can learn the graph embedding in a purely unsupervised way in an end-to-end framework. These algorithms exploit edge-level information in two ways. One simple way is to adopt an autoencoder framework where the encoder employs graph convolutional layers to embed the graph into the latent representation upon which a decoder is used to reconstruct the graph structure \cite{kipf2016variational,pan2018adversarially}.  Another popular way is to utilize the negative sampling approach which samples a portion of node pairs as negative pairs while existing node pairs with links in the graphs are positive pairs. Then a logistic regression layer is applied to distinguish between positive and negative pairs \cite{hamilton2017inductive}. 
    \end{itemize}
    
    In Table \ref{tab:summary_gcn}, we summarize the main characteristics of representative RecGNNs and ConvGNNs. Input sources, pooling layers, readout layers, and time complexity are compared among various models. In more detail, we only compare the time complexity of the message passing/graph convolution operation in each model. As methods in \cite{bruna2013spectral} and \cite{henaff2015deep} require eigenvalue decomposition, the time complexity is $O(n^3)$. The time complexity of \cite{tran2018filter} is also $O(n^3)$ due to the node pair-wise shortest path computation.  Other methods incur equivalent time complexity, which is $O(m)$ if the graph adjacency matrix is sparse and is $O(n^2)$ otherwise. This is because in these methods the computation of each node $v_i$'s representation involves its $d_i$ neighbors, and the sum of $d_i$ over all nodes exactly equals the number of edges. The time complexity of several methods are missing in Table \ref{tab:summary_gcn}. These methods either lack a time complexity analysis in their papers or report the time complexity of their overall models or algorithms. %We refer readers to Table \ref{tab:complex} which compares the time and memory complexity of several GNN training algorithms.

    %are mostly learning algorithms of graph neural networks. For time complexity analysis of a group of learning algorithms, please refer to Table \ref{tab:complex}.
    
    %list the details of the inputs and outputs of the main GCNs methods. In particular, we summarize output mechanisms in between each GCN layer and in the final layer of each method. The output mechanisms may involve several pooling operations, which are discussed in Section \ref{sec:pooling}.

    \begin{table*}[]
    	\caption{Summary of RecGNNs and ConvGNNs. %Notations are predefined in Table \ref{tab:notations}. 
    	Missing values (``-") in pooling and readout layers indicate that the method only experiments on node-level/edge-level tasks. 
    	%When both pooling and readout layers are unavailable, it means this method is only experimented with node-level tasks. For fair comparison, we only report the time complexity of computing the aggregation/convolution function. For simplicity, we assume $m$ is far greater than $n$ as well as other hyper-parameters such as the dimension of node hidden vectors.
    	}
		\label{tab:summary_gcn}
		\centering
        \begin{tabular}{l l l l l l }
        \toprule
        \multirow{2}{*}{Approach} & \multirow{2}{*}{Category} & \multirow{2}{*}{Inputs}  & \multirow{2}{*}{Pooling}  & \multirow{2}{*}{Readout} & \multirow{2}{*}{Time Complexity} \\ 
        &&&&& \\ \midrule
        GNN* (2009) \cite{scarselli2009graph}& RecGNN & $A,X,X^e$ & - &  a dummy super node          & $O(m)$ \\ \midrule
        GraphESN (2010) \cite{gallicchio2010graph} & RecGNN & $A,X$ & - & mean & $O(m)$\\ \midrule
        GGNN (2015) \cite{li2015gated} & RecGNN & $A,X$ & - & attention sum & $O(m)$ \\ \midrule
        SSE (2018) \cite{dai2018learning} & RecGNN & $A,X$ & - & - & - \\ \midrule
        Spectral CNN (2014) \cite{bruna2013spectral}&  Spectral-based ConvGNN &  $A,X$   & spectral clustering+max pooling & max & $O(n^3)$ \\ \midrule
        Henaff et al. (2015) \cite{henaff2015deep} & Spectral-based ConvGNN&  $A,X$   & spectral clustering+max pooling &  & $O(n^3)$ \\ \midrule
        ChebNet (2016) \cite{defferrard2016convolutional} & Spectral-based ConvGNN & $A,X$ &  efficient pooling & sum & $O(m)$ \\ \midrule
        GCN (2017) \cite{kipf2017semi} & Spectral-based ConvGNN & $A,X$ & - & - & $O(m)$ \\ \midrule
        CayleyNet (2017) \cite{levie2017cayleynets}& Spectral-based ConvGNN & $A,X$ & mean/graclus pooling & - & $O(m)$\\ \midrule
        AGCN (2018) \cite{li2018adaptive} & Spectral-based ConvGNN &$A,X$ & max pooling & sum & $O(n^2)$ \\ \midrule
        DualGCN (2018) \cite{zhuang2018dual} & Spectral-based ConvGNN & $A,X$ & - & - & $O(m)$\\ \midrule
        %SGC (2019) \cite{wu2019simplifying} & Spectral-based ConvGNN & $A,X$ & - & - & $O(n^2)$ \\ \midrule
        NN4G (2009) \cite{micheli2009neural} & Spatial-based ConvGNN & $A,X$ & - & sum/mean &$O(m)$ \\ \midrule
        DCNN (2016) \cite{atwood2016diffusion} & Spatial-based ConvGNN & $A,X$ & - & mean & $O(n^2)$ \\ \midrule
        PATCHY-SAN (2016) \cite{niepert2016learning} & Spatial-based ConvGNN & $A,X,X^e$  &  - & sum & -\\ \midrule
        MPNN (2017) \cite{gilmer2017neural} & Spatial-based ConvGNN & $A,X,X^e$ & - & attention sum/set2set & $O(m)$ \\ \midrule
        GraphSage (2017) \cite{hamilton2017inductive} & Spatial-based ConvGNN & $A,X$ & - & - & - \\ \midrule
        GAT (2017) \cite{velickovic2017graph} & Spatial-based ConvGNN & $A,X$ & - & - & $O(m)$\\ \midrule
        MoNet (2017) \cite{monti2017geometric} & Spatial-based ConvGNN & $A,X$ & - & - & $O(m)$  \\ \midrule
        LGCN (2018) \cite{gao2018large} & Spatial-based ConvGNN & $A,X$ & - & - & -\\ \midrule
        PGC-DGCNN (2018) \cite{tran2018filter} & Spatial-based ConvGNN & $A,X$ & sort pooling & attention sum & $O(n^3)$ \\ \midrule
        CGMM (2018) \cite{bacciu2018contextual} &Spatial-based ConvGNN & $A,X,X^e$ & - & sum & - \\ \midrule
        GAAN (2018) \cite{zhang2018gaan} & Spatial-based ConvGNN &$A,X$ & - & - & $O(m)$ \\ \midrule
        FastGCN (2018) \cite{chen2018fastgcn} & Spatial-based ConvGNN & $A,X$ & - & - & - \\ \midrule
        StoGCN (2018) \cite{chen2018stochastic} & Spatial-based ConvGNN & $A,X$ & - & - & - \\ \midrule
        Huang et al. (2018) \cite{huang2018ada} & Spatial-based ConvGNN & $A,X$ & - & - & - \\ \midrule
        DGCNN (2018) \cite{zhang2018end} & Spatial-based ConvGNN & $A,X$ & sort pooling &  - & $O(m)$ \\ \midrule
        DiffPool (2018) \cite{ying2018hierarchical}& Spatial-based ConvGNN & $A,X$ & differential pooling & mean & $O(n^2)$ \\ \midrule
        %Li et al. (2018) \cite{li2018deeper} & Spatial-based GCN & $A,X$ & - & - & - \\ \midrule
        GeniePath (2019) \cite{liu2018geniepath} & Spatial-based ConvGNN & $A,X$ & - & - & $O(m)$  \\ \midrule
        DGI (2019) \cite{velivckovic2019deep} & Spatial-based ConvGNN & $A,X$ & - & - & $O(m)$ \\ \midrule
        GIN (2019) \cite{xu2019how} & Spatial-based ConvGNN &$A,X$ & - & sum &  $O(m)$ \\ \midrule
        %CapsGNN (2019) \cite{xinyi2019capsule} & Spatial-based ConvGNN & $A,X$ & - & - & $O(m)$ \\ \midrule
        ClusterGCN (2019) \cite{chiang2019cluster} & Spatial-based ConvGNN & $A,X$ & - & - & - \\ \midrule
        %k-GNN (2019) \cite{morris2018weisfeiler} & Spatial-based GCN & & - & - & \\ \bottomrule
        \end{tabular}
        \end{table*}

    \section{Recurrent Graph Neural Networks}
    \label{sec:grn}
    Recurrent graph neural networks (RecGNNs) are mostly pioneer works of GNNs.  They apply the same set of parameters recurrently over nodes in a graph to extract high-level node representations. Constrained by computational power, earlier research mainly focused on directed acyclic graphs \cite{sperduti1997supervised,micheli2004contextual}.%hagenbuchner2003self

    Graph Neural Network (GNN*\footnote{As GNN is used to represent broad graph neural networks in the survey, we name this particular method GNN* to avoid ambiguity.}) proposed by Scarselli et al. extends prior recurrent models to handle general types of graphs, e.g., acyclic, cyclic, directed, and undirected graphs \cite{scarselli2009graph}.  Based on an information diffusion mechanism, GNN* updates nodes' states by exchanging neighborhood information recurrently until a stable equilibrium is reached.  A node's hidden state is recurrently updated by
    
    %\begin{equation}
    %\label{eq:gnn}
    %\mathbf{h}_v^{(t)} = f(\mathbf{x}_v,\{\mathbf{x^e}_{(v,u)}, %\mathbf{x}_{u}, \mathbf{h}^{(t-1)}_{u},  u\in N(v)\})
    %\end{equation}
    
    \begin{equation}
    \label{eq:gnn}
    \mathbf{h}_v^{(t)} = \sum_{u\in N(v)}f(\mathbf{x}_v,\mathbf{x^e}_{(v,u)}, \mathbf{x}_{u}, \mathbf{h}^{(t-1)}_{u}),
    \end{equation}
    where $f(\cdot)$ is a parametric function, and $\mathbf{h}^{(0)}_v$ is initialized randomly. The sum operation enables GNN* to be applicable to all nodes, even if the number of neighbors differs and no neighborhood ordering is known. 
    To ensure convergence, the recurrent function $f(\cdot)$ must be a contraction mapping, which shrinks the distance between two points after projecting them into a latent space. 
    In the case of $f(\cdot)$ being a neural network, a penalty term has to be imposed on the Jacobian matrix of parameters.  
    When a convergence criterion is satisfied,  the last step node hidden states are forwarded to a readout layer. GNN* alternates the stage of node state propagation and the stage of parameter gradient computation to minimize a training objective.  This strategy enables GNN* to handle cyclic graphs.  In follow-up works,  Graph Echo State Network (GraphESN) \cite{gallicchio2010graph} extends echo state networks to improve the training efficiency of GNN*. GraphESN consists of an encoder and an output layer. The encoder is randomly initialized and requires no training. It implements a contractive state transition function to recurrently update node states until the global graph state reaches convergence. Afterward, the output layer is trained by taking the fixed node states as inputs.

    %is responsible for deriving stable node states by 
    
    Gated Graph Neural Network (GGNN) \cite{li2015gated} employs a gated recurrent unit (GRU) \cite{cho2014learning} as a recurrent function, reducing the recurrence to a fixed number of steps. The advantage is that it no longer needs to constrain parameters to ensure convergence. A node hidden state is updated by its previous hidden states and its neighboring hidden states, defined as
    \begin{equation}
    \mathbf{h}_v^{(t)} = GRU(\mathbf{h}_v^{(t-1)},\sum_{u\in N(v)}\mathbf{W}\mathbf{h}_u^{(t-1)}),
    \end{equation}
    where $\mathbf{h}_v^{(0)}=\mathbf{x}_v$.
     Different from GNN* and GraphESN, GGNN uses the back-propagation through time (BPTT) algorithm to learn the model parameters.  This can be problematic for large graphs, as GGNN needs to run the recurrent function multiple times over all nodes, requiring the intermediate states of all nodes to be stored in memory.
    
    Stochastic Steady-state Embedding (SSE) proposes a learning algorithm that is more scalable to large graphs \cite{dai2018learning}. SSE updates node hidden states recurrently in a stochastic and asynchronous fashion.  It alternatively samples a batch of nodes for state update and a batch of nodes for gradient computation. To maintain stability, the recurrent function of SSE is defined as a weighted average of the historical states and new states, which takes the form
    
    \begin{equation}
    \label{eq:sse1}
    \mathbf{h}_v^{(t)} = (1-\alpha)\mathbf{h}_v^{(t-1)}+\alpha \mathbf{W_1}\sigma(\mathbf{W_2}[\mathbf{x}_v,\sum_{u\in N(v)}[\mathbf{h}_u^{(t-1)},\mathbf{x}_u]]),
    \end{equation}
    where $\alpha$ is a hyper-parameter, and $\mathbf{h}_v^{(0)}$ is initialized randomly.  While conceptually important, SSE does not theoretically prove that the node states will gradually converge to fixed points by applying Equation \ref{eq:sse1} repeatedly.

	\section{Convolutional Graph Neural  Networks}
	\label{sec:gcn}

\begin{figure}
		\centering
		\subfloat[Recurrent Graph Neural Networks (RecGNNs). RecGNNs use the same graph recurrent layer (Grec) in updating node representations. ]{\includegraphics[width=3.5in]{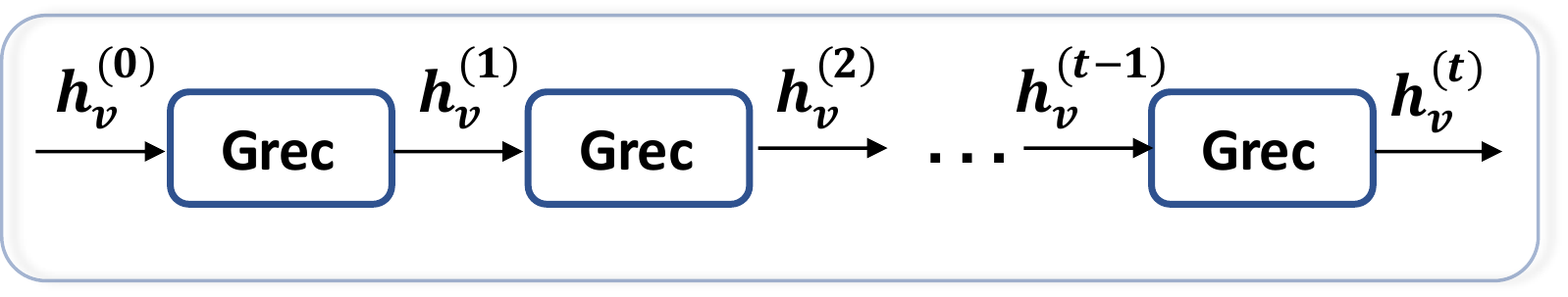}%
			\label{fig:grnt}}
		\hfill
		\subfloat[Convolutional Graph Neural Networks (ConvGNNs). ConvGNNs use a different graph convolutional layer (Gconv) in updating node representations. ]{\includegraphics[width=3.5in]{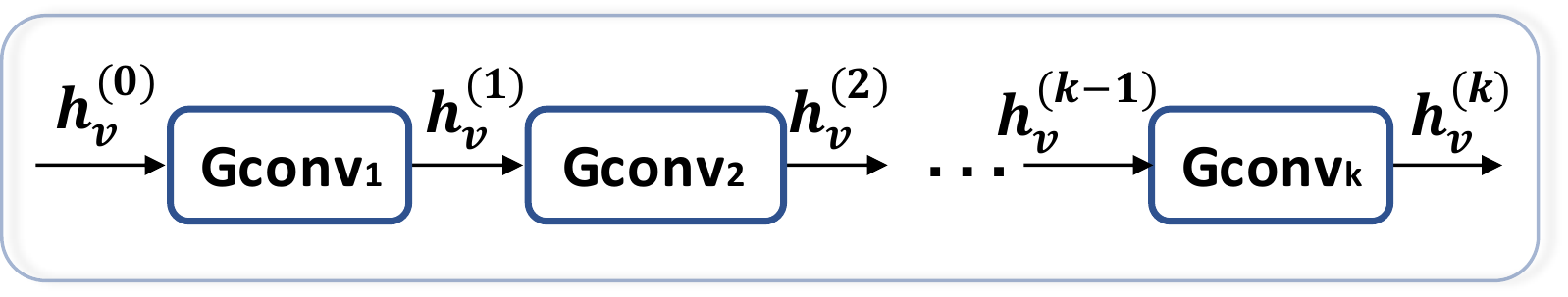}%
			\label{fig:gcnk}}
		\caption{RecGNNs v.s. ConvGNNs}
		\label{fig:spgcn}
	\end{figure}

    Convolutional graph neural networks (ConvGNNs) are closely related to recurrent graph  neural networks. Instead of iterating node states with contractive constraints, ConvGNNs address the cyclic mutual dependencies architecturally using a fixed number of layers with different weights in each layer. 
    This key distinction is illustrated in Figure \ref{fig:spgcn}. 
    As graph convolutions are more efficient and convenient to composite with other neural networks, the popularity of ConvGNNs has been rapidly growing in recent years. 
    %For example, graph pooling modules can be interleaved with graph convolutional layers, to coarsen graphs into high-level sub-structures. As shown in Fig \ref{fig:gcnpool}, such an architecture design can be used to extract graph-level representations and to perform graph classification tasks.
	ConvGNNs fall into two categories, spectral-based and spatial-based. Spectral-based approaches define graph convolutions by introducing filters from the perspective of graph signal processing \cite{shuman2013emerging} where the graph convolutional operation is interpreted as removing noises from graph signals. Spatial-based approaches inherit ideas from RecGNNs to define graph convolutions by information propagation. Since GCN \cite{kipf2017semi} bridged the gap between spectral-based approaches and spatial-based approaches, spatial-based methods have developed rapidly recently due to its attractive efficiency, flexibility, and generality.
	%In the following, we introduce spectral-based ConvGNNs, spatial-based ConvGNNs, and graph pooling modules sequentially. 

	\subsection{Spectral-based ConvGNNs}
	\label{sec:spectral_gcn}
	%Spectral-based methods have a solid foundation in graph signal processing \cite{shuman2013emerging}. We first give some basic knowledge background of graph signal processing, after which we review the representative research on the spectral-based GCNs.
	
	\vspace{2mm}
	\textbf{Background} Spectral-based methods have a solid mathematical foundation in graph signal processing \cite{shuman2013emerging,sandryhaila2013discrete,chen2015discrete}. They assume graphs to be undirected. The normalized graph Laplacian matrix is a mathematical representation of an undirected graph, defined as $\mathbf{L}=\mathbf{I_n}- \mathbf{D}^{-\frac{1}{2}}\mathbf{A}\mathbf{D}^{-\frac{1}{2}}$, where $\mathbf{D}$ is a diagonal matrix of node degrees, $\mathbf{D}_{ii}=\sum_j(\mathbf{A}_{i,j})$. The normalized graph Laplacian matrix possesses the property of being real symmetric positive semidefinite.  With this property, the normalized Laplacian matrix can be factored as $\mathbf{L}=\mathbf{U}\mathbf{\Lambda}\mathbf{U}^T$, where $\mathbf{U}=[\mathbf{u_0},\mathbf{u_1},\cdots,\mathbf{u_{n-1}}] \in \mathbf{R}^{n\times n}$ is the matrix of eigenvectors ordered by eigenvalues and $\mathbf{\Lambda}$ is the diagonal matrix of eigenvalues (spectrum), $\mathbf{\Lambda}_{ii}=\lambda_i$. The eigenvectors of the normalized Laplacian matrix form an orthonormal space, in mathematical words $\mathbf{U}^T\mathbf{U}=\mathbf{I}$.  In graph signal processing, a graph signal $\mathbf{x} \in  \mathbf{R}^n$ is a feature vector of all nodes of a graph where $x_i$ is the value of the $i^{th}$ node. The \textit{graph Fourier transform} to a signal $\mathbf{x}$ is defined as $\mathscr{F}(\mathbf{x}) = \mathbf{U}^T\mathbf{x}$, and the inverse graph Fourier transform  is defined as $\mathscr{F}^{-1}(\mathbf{\hat{x}}) = \mathbf{U}\hat{\mathbf{x}}$, where $\mathbf{\hat{x}}$ represents the resulted signal from the graph Fourier transform. The graph Fourier transform projects the input graph signal to the orthonormal space where the basis is formed by eigenvectors of the normalized graph Laplacian. Elements of the transformed signal $\mathbf{\hat{x}}$  are the coordinates of the graph signal in the new space so that the input signal can be represented as $\mathbf{x}=\sum_i \hat{x}_i\mathbf{u}_i$, which is exactly the inverse graph Fourier transform. Now the graph convolution of the input signal $\mathbf{x}$ with a filter $\mathbf{g}\in \mathbf{R}^n$ is defined as 
	\begin{equation}
	\label{eq:graphconv}
	\begin{aligned}
	\mathbf{x} \ast_G \mathbf{g}&= \mathscr{F}^{-1}(\mathscr{F}(\mathbf{x})\odot\mathscr{F}(\mathbf{g})) \\
	&  = \mathbf{U}(\mathbf{U}^T\mathbf{x}\odot \mathbf{U}^T \mathbf{g}),
	\end{aligned}
	\end{equation}
	where $\odot$ denotes the element-wise product.  If we denote a filter as $\mathbf{g_\theta} = diag(\mathbf{U}^T\mathbf{g})$, then the spectral graph convolution is simplified as 
	\begin{equation}
	\label{eq:2}
	\mathbf{x} \ast_G \mathbf{g_\theta} = \mathbf{U}\mathbf{g_\theta} \mathbf{U}^T\mathbf{x}.
	\end{equation}
	%The following works all comply with this definition.
	Spectral-based ConvGNNs all follow this definition. The key difference lies in the choice of the filter $\mathbf{g_\theta}$.
	
	Spectral Convolutional Neural Network (Spectral CNN) \cite{bruna2013spectral} assumes the filter $\mathbf{g_\theta}=\mathbf{\Theta}_{i,j}^{(k)}$ is a set of learnable parameters and considers graph signals with multiple channels. The graph convolutional layer of Spectral CNN is defined as 
	\begin{equation}
	\label{eq:3}
	\mathbf{H}_{:,j}^{(k)} = \sigma(\sum_{i=1}^{f_{k-1}}\mathbf{U}\mathbf{\Theta}_{i,j}^{(k)}\mathbf{U}^T\mathbf{H}_{:,i}^{(k-1)}) \quad(j=1,2,\cdots, f_k),
	\end{equation}
	where $k$ is the layer index, $\mathbf{H}^{(k-1)} \in \mathbf{R}^{n\times f_{k-1}}$ is the input graph signal, $\mathbf{H}^{(0)}=\mathbf{X}$, $f_{k-1}$ is the number of input channels and $f_k$ is the number of  output channels, $\mathbf{\Theta}_{i,j}^{(k)}$ is a diagonal matrix filled with learnable parameters.  Due to the eigen-decomposition of the Laplacian matrix, Spectral CNN faces three limitations. First, any perturbation to a graph results in a change of eigenbasis. Second, the learned filters are domain dependent, meaning they cannot be applied to a graph with a different structure. Third, eigen-decomposition requires $O(n^3)$ computational complexity.  In follow-up works, ChebNet \cite{defferrard2016convolutional} and GCN \cite{kipf2017semi} reduce the computational complexity to $O(m)$ by making several approximations and simplifications. 
	
	Chebyshev Spectral CNN (ChebNet) \cite{defferrard2016convolutional} approximates the filter $\mathbf{g}_\theta$ by Chebyshev polynomials of the diagonal matrix of eigenvalues, i.e, $\mathbf{g_\theta}= \sum_{i=0}^{K} \theta_i T_i(\mathbf{\tilde{\Lambda}})$, where $\mathbf{\tilde{\Lambda}} = 2\mathbf{\Lambda}/\lambda_{max}-\mathbf{I_n}$, and the values of $\mathbf{\tilde{\Lambda}}$ lie in $[-1,1]$. The Chebyshev polynomials are defined recursively by $T_i(\mathbf{x}) = 2\mathbf{x}T_{i-1}(\mathbf{x})-T_{i-2}(\mathbf{x})$ with $T_0(\mathbf{x}) = 1$ and $T_1(\mathbf{x})=\mathbf{x}$. As a result, the convolution of a graph signal $\mathbf{x}$ with the defined filter $\mathbf{g_\theta}$ is 
	\begin{equation}
	\label{eq:chebpoly}
	\mathbf{x} \ast_G \mathbf{g_\theta} = \mathbf{U}(\sum_{i=0}^{K} \theta_i T_i(\mathbf{\tilde{\Lambda}}))\mathbf{U}^T\mathbf{x}, \\
	\end{equation}
	where $\mathbf{\tilde{L}} = 2\mathbf{L}/\lambda_{max}-\mathbf{I_n}$. 
	As $T_i(\mathbf{\tilde{L}})=\mathbf{U}T_i(\mathbf{\tilde{\Lambda}})\mathbf{U}^T$, which can be proven by induction on $i$, ChebNet takes the form, 
	\begin{equation}
	\label{eq:chebnet}
	\mathbf{x} \ast_G \mathbf{g_\theta} = \sum_{i=0}^{K} \theta_iT_i(\mathbf{\tilde{L}})\mathbf{x}, 
	\end{equation}
	As an improvement over Spectral CNN, the filters defined by ChebNet are localized in space, which means filters can extract local features independently of the graph size. The spectrum of ChebNet is mapped to $[-1,1]$ linearly. CayleyNet \cite{levie2017cayleynets} further applies Cayley polynomials which are parametric rational complex functions to capture narrow frequency bands. The spectral graph convolution of CayleyNet is defined as 
	\begin{equation}
	    \mathbf{x} \ast_G \mathbf{g_\theta} = c_0\mathbf{x}+2Re\{\sum_{j=1}^rc_j(h\mathbf{L}- i\mathbf{I})^j(h\mathbf{L}+ i\mathbf{I})^{-j}\mathbf{x}\},
	\end{equation}
    where $Re(\cdot)$ returns the real part of a complex number, $c_0$ is a real coefficent, $c_j$ is a complex coefficent, $i$ is the imaginary number, and $h$ is a parameter which controls the spectrum of a Cayley filter. While preserving spatial locality, CayleyNet shows that ChebNet can be considered as a special case of CayleyNet.
	%From Equation \ref{eq:chebnet}, ChebNet implicitly avoids the computation of the graph Fourier basis, reducing the computation complexity from $O(N^3)$ to $O(KM)$.  Since $T_i(\mathbf{\tilde{L}})$ is a polynomial of $\mathbf{\tilde{L}}$ of $i^{th}$ order, $T_i(\mathbf{\tilde{L}})\mathbf{x}$ operates locally on each node. Therefore, the filters of ChebNet are localized in space.

    Graph Convolutional Network (GCN) \cite{kipf2017semi} introduces a first-order approximation of ChebNet.  Assuming $K=1$ and $\lambda_{max} = 2$
	, Equation \ref{eq:chebnet} is simplified as 
	\begin{equation}
	\label{eq:1stchebnet}
	\mathbf{x} \ast_G \mathbf{g_\theta}= \theta_0\mathbf{x}-\theta_1\mathbf{D}^{-\frac{1}{2}}\mathbf{A}\mathbf{D}^{-\frac{1}{2}}\mathbf{x}.
	\end{equation}
%\footnote{Due to its impressive performance in many node classification tasks, 1stChebNet is simply termed as \textbf{GCN} and is considered as a strong baseline in the research community.})} 
	To restrain the number of parameters and avoid over-fitting, GCN further assume $\theta=\theta_0=-\theta_1$,  leading to the following definition of a graph convolution,
	\begin{equation}
	\label{eq:1stchebnetb}
	\mathbf{x} \ast_G \mathbf{g_\theta} = \theta(\mathbf{I_n}+\mathbf{D}^{-\frac{1}{2}}\mathbf{A}\mathbf{D}^{-\frac{1}{2}})\mathbf{x}.
	\end{equation}
	To allow multi-channels of inputs and outputs, GCN modifies Equation \ref{eq:1stchebnetb} into a compositional layer, defined as 
	\begin{equation}
	\label{eq:1stchebnetc}
	\mathbf{H}= \mathbf{X}\ast_G \mathbf{g_\Theta} = f(\mathbf{\bar{A}}\mathbf{X} \mathbf{\Theta}),
	\end{equation} 
	where $\mathbf{\bar{A}} =\mathbf{I_n}+\mathbf{D}^{-\frac{1}{2}}\mathbf{A}\mathbf{D}^{-\frac{1}{2}}$ and $f(\cdot)$ is an activation function.  Using $\mathbf{I_n}+\mathbf{D}^{-\frac{1}{2}}\mathbf{A}\mathbf{D}^{-\frac{1}{2}}$ empirically causes numerical instability to GCN. To address this problem, GCN applies a normalization trick to replace $\mathbf{\bar{A}}=\mathbf{I_n}+\mathbf{D}^{-\frac{1}{2}}\mathbf{A}\mathbf{D}^{-\frac{1}{2}}$ by $\mathbf{\bar{A}}=\tilde{\mathbf{D}}^{-\frac{1}{2}}\tilde{\mathbf{A}}\tilde{\mathbf{D}}^{-\frac{1}{2}}$ with $\tilde{\mathbf{A}}=\mathbf{A}+\mathbf{I_n}$ and $\tilde{\mathbf{D}}_{ii}=\sum_j \tilde{\mathbf{A}}_{ij}$. 
	%To incorporate structural information from local to global, one can stack multiple layers. GCN adopts a two-layer architecture for the node classification problem. The model is constructed as 
	%\begin{equation}
	    %\mathbf{Y} = %softmax(\mathbf{\bar{A}}f(\mathbf{\bar{A}}\mathbf{X} %\mathbf{\Theta}_1)\mathbf{\Theta}_2).
	%\end{equation}
	Being a spectral-based method, GCN can be also interpreted as a spatial-based method. 
	%GCN bridges the gap between spectral-based approaches and spatial-based approaches. 
	From a spatial-based perspective, GCN can be considered as aggregating feature information from a node's neighborhood. 
	%each row of the output $\mathbf{H}$ represents the latent representation of each node obtained by a linear transformation of aggregated information from the node itself and its neighboring nodes with weights specified by the row of $\mathbf{\bar{A}}$.  
	Equation \ref{eq:1stchebnetc} can be expressed as
	\begin{equation}
	    \mathbf{h}_v = f(\mathbf{\Theta}^T(\sum_{u\in\{N(v)\cup v\}} \bar{A}_{v,u}\mathbf{x}_u))\quad \forall v\in V.
	    \label{eq:sgcn}
	\end{equation}

    Several recent works made incremental improvements over GCN \cite{kipf2017semi} by exploring alternative symmetric matrices.
    %or simplifying multiple layers of graph convolution based on Equation \ref{eq:1stchebnetc}.
    Adaptive Graph Convolutional Network (AGCN) \cite{li2018adaptive} learns hidden structural relations unspecified by the graph adjacency matrix. It constructs a so-called residual graph adjacency matrix through a learnable distance function which takes two nodes' features as inputs. Dual Graph Convolutional Network (DGCN) \cite{zhuang2018dual} introduces a dual graph convolutional architecture with two graph convolutional layers in parallel. While these two layers share parameters, they use the normalized adjacency matrix $\mathbf{\bar{A}}$ and the positive pointwise mutual information (PPMI) matrix which captures nodes co-occurrence information through random walks sampled from a graph. 
    The PPMI matrix is defined as 
	\begin{equation}
	\mathbf{PPMI}_{v_1,v_2} = max(\log (\frac{count(v_1,v_2)\cdot |D|}{count(v_1)count(v_2)}),0),
	\end{equation}
	where $v_1,v_2\in V$, $|D|=\sum_{v_1,v_2}count(v_1,v_2)$ and the $count(\cdot)$ function returns the frequency that node $v$ and/or node $u$ co-occur/occur in sampled random walks. By ensembling outputs from dual graph convolutional layers, DGCN encodes both local and global structural information without the need to stack multiple graph convolutional layers. 
    %Simple Graph Convolution (SGC) \cite{wu2019simplifying} collapses a GCN with multiple layers into a linear equation by assuming identity activation functions between graph convolutional layers. 
    
    %\begin{comment}
    %The graph convolution of SGC takes the resulting form
    %\begin{equation}
     %   \mathbf{H}= \mathbf{\bar{A}}^K\mathbf{X}\mathbf{\Theta}.
    %\end{equation}
    %By taking powers $K>1$, SGC functions as a low-pass filter that produces smooth node representations.  
    %\end{comment}

	%As a result, Equation \ref{eq:sgcn} can be extended to process directed graphs with $\mathbf{\bar{A}}= \tilde{\mathbf{D}}^{-1}\tilde{\mathbf{A}}$.

	%\vspace{2mm}\noindent\textbf{Adaptive Graph Convolutional Network (AGCN).} To explore hidden structural relations unspecified by the graph Laplacian matrix,  Li et al. \cite{li2018adaptive} propose the adaptive graph convolutional network (AGCN). AGCN augments a graph with a so-called residual graph, which is constructed by computing a pairwise distance of nodes.  Despite being able to capture complement relational information, AGCN incurs expensive $O(N^2)$ computation.

    %However, a common drawback of spectral methods is they need to load the whole graph into the memory to perform graph convolution, which is not efficient in handling big graphs. 

    \subsection{Spatial-based ConvGNNs}
    \label{sec:spatial_gcn}
    Analogous to the convolutional operation of a conventional CNN on an image, spatial-based methods define graph convolutions based on a node's spatial relations. Images can be considered as a special form of graph with each pixel representing a node. 
    Each pixel is directly connected to its nearby pixels, as illustrated in Figure \ref{fig_first_case}.  A filter is applied to a $3\times 3$ patch by taking the weighted average of pixel values of the central node and its neighbors across each channel. Similarly, the spatial-based graph convolutions convolve the central node's representation with its neighbors' representations to derive the updated representation for the central node, as illustrated in Figure \ref{fig_second_case}. From another perspective, spatial-based ConvGNNs share the same idea of information propagation/message passing with RecGNNs. The spatial graph convolutional operation essentially propagates node information along edges. 
    %However, there are two differences between an image graph and a general graph. First, a node from an image graph has a fixed number of neighbors while a node from a general graph has a variant number of neighbors. Having a different size of neighborhood for each node impedes the efficiency of performing graph convolution. Spatial-based ConvGNNs may address this problem by sampling a fixed size of neighborhood such as \cite{niepert2016learning,hamilton2017inductive,gao2018large}.Second, the neighbors of a node from an image graph have relative positions/orders to the central node while the neighbors of a node from a general graph do not preserve this information. Due to lack of node relative positions/orders, the trainable weights/parameters of a graph filter are hard to be shared across different locations in a graph. Towards this issue, a standard way is to use node ordering invariant aggregation functions such as mean and sum, which assign equal weight to each neighbor of a node.To enable weight sharing across different locations, there are a number of methods which considers assigning different weights to a node's neighbors, including taking weights from a normalized adjacency matrix \cite{kipf2017semi,atwood2016diffusion,tran2018filter, li2018diffusion,yan2018spatial}, utilizing attention mechanisms \cite{velickovic2017graph,zhang2018gaan}, leveraging node pseudo coordinates \cite{monti2017geometric}, and choosing a nodes' top-k neighbors and applying standard 1D convolutions \cite{niepert2016learning,gao2018large}. 
   
    Neural Network for Graphs (NN4G) \cite{micheli2009neural}, proposed in parallel with GNN*, is the first work towards spatial-based ConvGNNs. Distinctively different from RecGNNs, NN4G learns graph mutual dependency through a compositional neural architecture with independent parameters at each layer. The neighborhood of a node can be extended through incremental construction of the architecture.
    %which was proposed at the same time when Scarselli et al. introduced the recurrent graph nerual network GNN \cite{scarselli2009graph}.Unlike GNN, NN4G is free of constractive constraints. NN4G learns graph mutual dependency through a compositional neural architecture with independent parameters at each layer.  It shows the receptive field of a node can be incrementally extended through the compositionality of multiple layers. 
    NN4G performs graph convolutions by summing up a node's neighborhood information directly. It also applies residual connections and skip connections to memorize information over each layer. As a result, NN4G derives its next layer node states by
    \begin{equation}
        \mathbf{h}_v^{(k)} = f(\mathbf{W}^{(k)^T}\mathbf{x}_v+\sum_{i=1}^{k-1}\sum_{u\in N(v)}\mathbf{\Theta}^{(k)^T}\mathbf{h}_u^{(k-1)}),
        \label{eq:nn4g}
    \end{equation}
    where $f(\cdot)$ is an activation function and $\mathbf{h}_v^{(0)}=\mathbf{0}$.  Equation \ref{eq:nn4g} can also be written in a matrix form:
    \begin{equation}
        \mathbf{H}^{(k)} = f(\mathbf{X}\mathbf{W}^{(k)}+\sum_{i=1}^{k-1}\mathbf{A}\mathbf{H}^{(k-1)}\mathbf{\Theta}^{(k)}),
    \end{equation}
    which resembles the form of GCN \cite{kipf2017semi}. One difference is that NN4G uses the unnormalized adjacency matrix which may potentially cause hidden node states to have extremely different scales.
    Contextual Graph Markov Model (CGMM) \cite{bacciu2018contextual} proposes a probabilistic model inspired by NN4G. While maintaining spatial locality, CGMM has the benefit of probabilistic interpretability.

    Diffusion Convolutional Neural Network (DCNN) \cite{atwood2016diffusion} regards graph convolutions as a diffusion process. It assumes information is transferred from one node to one of its neighboring nodes with a certain transition probability so that information distribution can reach equilibrium after several rounds. DCNN defines the diffusion graph convolution as
    
    \begin{equation}
	\label{eq:dcnn}
	\mathbf{H}^{(k)} = f(\mathbf{W}^{(k)}\odot \mathbf{P}^{k} \mathbf{X}),
	\end{equation}
    where $f(\cdot)$ is an activation function and the probability transition matrix $\mathbf{P}\in\mathbf{R}^{n\times n}$ is computed by $\mathbf{P}= \mathbf{D}^{-1}\mathbf{A}$.  Note that in DCNN, the hidden representation matrix $\mathbf{H}^{(k)}$ remains the same dimension as the input feature matrix $\mathbf{X}$ and is not a function of its previous hidden representation matrix $\mathbf{H}^{(k-1)}$. DCNN concatenates $\mathbf{H}^{(1)},\mathbf{H}^{(2)},\cdots,\mathbf{H}^{( K)} $ together as the final model outputs. As the stationary distribution of a diffusion process is a summation of power series of probability transition matrices, Diffusion Graph Convolution (DGC) \cite{li2018diffusion} sums up outputs at each diffusion step instead of concatenation. It defines the diffusion graph convolution by
    \begin{equation}
    \label{eq:dcn}
    \mathbf{H} = \sum_{k=0}^{K} f(\mathbf{P}^k\mathbf{X}\mathbf{W}^{(k)}),
\end{equation}
    where $\mathbf{W}^{(k)}\in \mathbf{R}^{D\times F}$ and $f(\cdot)$ is an activation function. 
    Using the power of a transition probability matrix implies that distant neighbors contribute very little information to a central node. PGC-DGCNN \cite{tran2018filter} increases the contributions of distant neighbors based on shortest paths.  It defines a shortest path adjacency matrix $\mathbf{S}^{(j)}$. If the shortest path from a node $v$ to a node $u$ is of length $j$, then $\mathbf{S}^{(j)}_{v,u}=1$ otherwise 0. With a hyperparameter $r$ to control the receptive field size, PGC-DGCNN introduces a graph convolutional operation as follows 
    \begin{equation}
        \mathbf{H}^{(k)} = \parallel_{j=0}^r f((\tilde{\mathbf{D}}^{(j)})^{-1}\mathbf{S}^{(j)}\mathbf{H}^{(k-1)}\mathbf{W}^{(j,k)}),
    \end{equation}
    where $\tilde{D}^{(j)}_{ii}=\sum_l S_{i,l}^{(j)}$, $\mathbf{H}^{(0)}=\mathbf{X}$, and $\parallel$ represents the concatenation of vectors. The calculation of the shortest path adjacency matrix can be expensive with $O(n^3)$ at maximum.  Partition Graph Convolution (PGC) \cite{yan2018spatial} partitions a node's neighbors into $Q$ groups based on certain criteria not limited to shortest paths.  PGC constructs $Q$ adjacency matrices according to the defined neighborhood by each group. Then, PGC applies GCN \cite{kipf2017semi} with a different parameter matrix to each neighbor group and sums the results:
	\begin{equation}
	\label{eq:pgcn}
	\mathbf{H}^{(k)} = \sum_{j=1}^{Q} \mathbf{\bar{A}}^{(j)}\mathbf{H}^{(k-1)}\mathbf{W}^{(j,k)},
	\end{equation}
	where $\mathbf{H}^{(0)}=\mathbf{X}$, $\mathbf{\bar{A}}^{(j)}=\tilde{(\mathbf{D}}^{(j)})^{-\frac{1}{2}}\tilde{\mathbf{A}}^{(j)}\tilde{(\mathbf{D}}^{(j)})^{-\frac{1}{2}}$ and $\tilde{\mathbf{A}}^{(j)}=\mathbf{A}^{(j)}+\mathbf{I}$.
	
    %As weights of graph filters of DCNN and DGC are determined by transition probabilities, the graph convolution of DCNN and DGC are invariant to node orderings. However, the complexity of computing power series of probability transition matrices is $O(Kn^2)$. Meanwhile,  both DCNN and DGC need to load the whole graph into memory for processing graph convolution. Therefore they are not scalable to large graphs.

    Message Passing Neural Network (MPNN) \cite{gilmer2017neural} outlines a general framework of spatial-based ConvGNNs. It treats graph convolutions as a message passing process in which information can be passed from one node to another along edges directly. MPNN runs K-step message passing iterations to let information propagate further. The message passing function (namely the spatial graph convolution) is defined as
	\begin{equation}
	\mathbf{h}_v^{(k)} = U_k(\mathbf{h}_v^{(k-1)},\sum_{u\in N(v)} M_k(\mathbf{h}_v^{(k-1)},\mathbf{h}_u^{(k-1)},\mathbf{x}^e_{vu})),
	\end{equation}
	where $\mathbf{h}_v^{(0)}=\mathbf{x}_v$, $U_k(\cdot)$ and $M_k(\cdot)$ are functions with learnable parameters.  After deriving the hidden representations of each node, $\mathbf{h}_v^{(K)}$ can be passed to an output layer to perform node-level prediction tasks or to a readout function to perform graph-level prediction tasks.  The readout function generates a representation of the entire graph based on node hidden representations. It is generally defined as
	\begin{equation}
	\mathbf{h}_G = R({\mathbf{h}_v^{(K)}|v\in G}),
	\end{equation}
	where $R(\cdot)$ represents the readout function with learnable parameters. MPNN can cover many existing GNNs by assuming different forms of $U_k(\cdot),M_k(\cdot)$, and $R(\cdot)$, such as \cite{kipf2017semi, duvenaud2015convolutional,kearnes2016molecular,schutt2017quantum}. However, Graph Isomorphism Network (GIN) \cite{xu2019how} finds that previous MPNN-based methods are incapable of distinguishing different graph structures based on the graph embedding they produced. To amend this drawback,  GIN adjusts the weight of the central node by a learnable parameter $\epsilon^{(k)}$. It performs graph convolutions by
	\begin{equation}
	    \mathbf{h}_v^{(k)} = MLP((1+\epsilon^{(k)})\mathbf{h}_v^{(k-1)}+\sum_{u\in N(v)}\mathbf{h}_u^{(k-1)}),
	\end{equation}
    where $MLP(\cdot)$ represents a multi-layer perceptron.
    
	%the training scheme of MPNNs may be inefficient. In order to improve computation efficiency, MPNNs split the node hidden representation $\mathbf{h}_v^k$ into $s$ parts and apply the spatial convolution on each small part separately. However, MPNN still has to store all intermediate hidden states into memory for backpropagation. 
    As the number of neighbors of a node can vary from one to a thousand or even more, it is inefficient to take the full size of a node's neighborhood. GraphSage \cite{hamilton2017inductive} adopts sampling to obtain a fixed number of neighbors for each node. It performs graph convolutions by 
	\begin{equation}
	\mathbf{h}^{(k)}_v=\sigma(\mathbf{W}^{(k)}\cdot f_k(\mathbf{h}_v^{(k-1)},\{\mathbf{h}_u^{(k-1)},\forall u \in S_{\mathcal{N}(v)}\})),
	\end{equation}
	where $\mathbf{h}^{(0)}_v=\mathbf{x}_v$, $f_k(\cdot)$ is an aggregation function, $S_{\mathcal{N}(v)}$ is a random sample of the node $v$'s neighbors. The aggregation function should be invariant to the permutations of node orderings such as a mean, sum or max function. %However, GraphSage empirically finds that an LSTM \cite{hochreiter1997long} aggregator which takes a random sequence of a node's neighbors performs marginally better than a mean aggregator. This implicitly suggests that the mean aggregator may lack expressive capability. 

	%attention means normalization, i.e. there exist some coefficients e_{i, j} which are computed e.g. via a neural network layer from h_i and h_j and which measure the connective strength from node j to node i.

	\begin{figure}[]
		\centering
		\subfloat[GCN \cite{kipf2017semi} explicitly assigns a non-parametric weight $a_{ij}=\frac{1}{\sqrt{deg(v_i)deg(v_j)}}$ to the neighbor $v_j$ of $v_i$ during the aggregation process. ]{\includegraphics[width=1.5in]{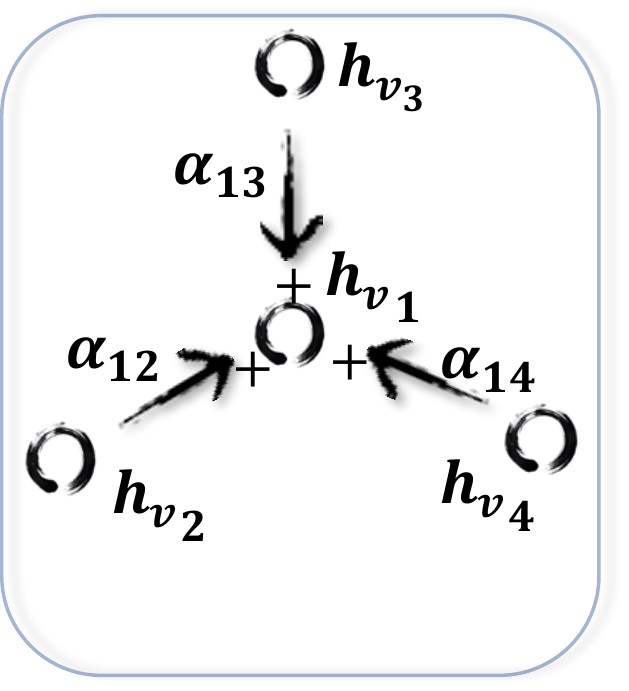}%
			\label{fig:demo_gcn}}
		\hfill
		\subfloat[GAT \cite{velickovic2017graph} implicitly captures the weight $a_{ij}$ via an end-to-end neural network architecture, so that more important nodes receive larger weights. ]{\includegraphics[width=1.5in]{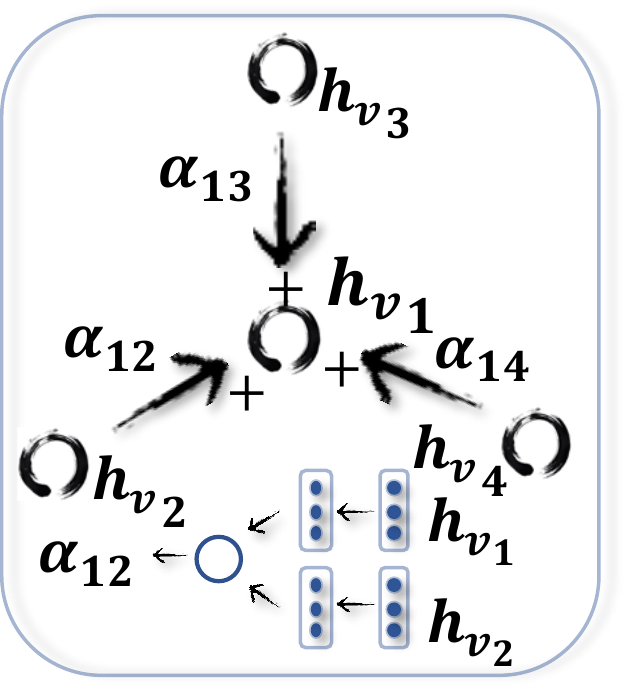}%
			\label{fig:demo_gat}}
		\caption{Differences between GCN \cite{kipf2017semi} and GAT \cite{velickovic2017graph}}
		\label{gcn_gat}
	\end{figure}

	Graph Attention Network (GAT) \cite{velickovic2017graph}
	assumes contributions of neighboring nodes to the central node are neither identical like GraphSage \cite{hamilton2017inductive}, nor pre-determined like GCN \cite{kipf2017semi} (this difference is illustrated in Figure \ref{gcn_gat}). GAT adopts attention mechanisms to learn the relative weights between two connected nodes.  The graph convolutional operation according to GAT is defined as,
	\begin{equation}
	\label{eq:13}
	\mathbf{h}_v^{(k)} = \sigma(\sum_{u\in\mathcal{N}(v)\cup v}\alpha_{vu}^{(k)}\mathbf{W}^{(k)}\mathbf{h}_u^{(k-1)}),
	\end{equation}
	where $\mathbf{h}^{(0)}_v=\mathbf{x}_v$. The attention weight $\alpha_{vu}^{(k)}$ measures the connective strength between the node $v$ and its neighbor $u$:
	\begin{equation}
	    \alpha_{vu}^{(k)} = softmax(g(\mathbf{a}^T[\mathbf{W}^{(k)}\mathbf{h}_v^{(k-1)}||\mathbf{W}^{(k)}\mathbf{h}_u^{(k-1)})),
	\end{equation}
	where $g(\cdot)$ is a LeakyReLU activation function and $\mathbf{a}$ is a vector of learnable parameters. The softmax function ensures that the attention weights sum up to one over all neighbors of the node $v$.
	GAT further performs the multi-head attention to increase the model's expressive capability. This shows an impressive improvement over GraphSage on node classification tasks. While GAT assumes the contributions of attention heads are equal, Gated  Attention  Network (GAAN) \cite{zhang2018gaan} introduces a self-attention  mechanism  which computes an additional attention score for each attention head. Apart from applying graph attention spatially, GeniePath \cite{liu2018geniepath} further proposes an LSTM-like gating mechanism to control information flow across graph convolutional layers. There are other graph attention models which might be of interest \cite{lee2018graph,abu2018watch}. However, they do not belong to the ConvGNN framework. %The introduction of graph attention comes at the cost of sacrificing model efficiency. GeniePath also proposes an implementation trick which reduces the computation complexity of Equation \ref{eq:13} from $O(n^2)$ to $O(m)$.
	
	Mixture Model Network (MoNet) \cite{monti2017geometric}
    adopts a different approach to assign different weights to a node's neighbors. It introduces node pseudo-coordinates to determine the relative position between a node and its neighbor. Once the relative position between two nodes is known, a weight function maps the relative position to the relative weight between these two nodes. In such a way, the parameters of a graph filter can be shared across different locations.  Under the MoNet framework, several existing approaches for manifolds such as Geodesic CNN (GCNN) \cite{masci2015geodesic}, Anisotropic CNN (ACNN) \cite{boscaini2016learning}, Spline CNN \cite{fey2018splinecnn}, and for graphs such as GCN \cite{kipf2017semi}, DCNN \cite{atwood2016diffusion} can be generalized as special instances of MoNet by constructing nonparametric weight functions.  MoNet additionally proposes a Gaussian kernel with learnable parameters to learn the weight function adaptively.
    
    Another distinct line of works achieve weight sharing across different locations by ranking a node's neighbors based on certain criteria and associating each ranking with a learnable weight. PATCHY-SAN \cite{niepert2016learning} orders neighbors of each node according to their graph labelings and selects the top $q$ neighbors.  Graph labelings are essentially node scores which can be derived by node degree, centrality, and Weisfeiler-Lehman color \cite{weisfeiler1968reduction,douglas2011weisfeiler}.  As each node now has a fixed number of ordered neighbors, graph-structured data can be converted into grid-structured data. PATCHY-SAN applies a standard 1D convolutional filter to aggregate neighborhood feature information where the order of the filter's weights corresponds to the order of a node's neighbors.
    The ranking criterion of PATCHY-SAN only considers graph structures, which requires heavy computation for data processing.  Large-scale Graph Convolutional Network (LGCN) \cite{gao2018large} ranks a node's neighbors based on node feature information. For each node, LGCN assembles a feature matrix which consists of its neighborhood and sorts this feature matrix along each column. The first $q$ rows of the sorted feature matrix are taken as the input data for the central node. %Similarly LGCN applies a standard 1D convolutional filter to aggregate neighborhood feature information. 

    %to do: To suit the scenario of large-scale graphs, LGCN proposes a subgraph training strategy, which puts the sampled subgraphs into a mini-batch.
    
	%Spatial-based methods define graph convolutions via aggregating feature information from neighbors. However, it is inefficient to store all the intermediate states into memory. To address this issue, several training strategies have been proposed, including sub-graph training for composition-based approaches such as GraphSage \cite{hamilton2017inductive}.  In addition, recent advances in spatial-based approaches tend to construct more complex network architectures. For examples, DualGCN \cite{zhuang2018dual} devises two graph convolutional networks to jointly embed the local consistency and the global consistency knowledge of a graph.  Tran et al. \cite{tran2018filter}  introduce a hyper-parameter to influence the receptive field size of a node.
	
\begin{table*}[]
\centering
\caption{Time and memory complexity comparison for ConvGNN training algorithms (summarized by \cite{chiang2019cluster}). $n$ is the total number of nodes. $m$ is the total number of edges. $K$ is the number of layers. $s$ is the batch size. $r$ is the number of neighbors being sampled for each node. For simplicity, the dimensions of the node hidden features remain constant, denoted by $d$.}
\label{tab:complex}
\begin{tabular}{llllll}
\toprule
     Complexity             & GCN \cite{kipf2017semi} & GraphSage \cite{hamilton2017inductive} & FastGCN \cite{chen2018fastgcn} & StoGCN \cite{chen2018stochastic}  & Cluster-GCN \cite{chiang2019cluster} \\ \midrule
Time    &  $O(Kmd+Knd^2)$    &   $O(r^Knd^2)$       &  $O(Krnd^2)$       &   $O(Kmd+Knd^2+r^Knd^2)$              &       $O(Kmd+Knd^2)$      \\ \hline
Memory &  $O(Knd+Kd^2)$   &    $O(sr^Kd+Kd^2)$       &   $O(Ksrd+Kd^2)$      & $O(Knd+Kd^2)$                 &  $O(Ksd+Kd^2)$    
\\ \bottomrule
\end{tabular}
\end{table*}
	
	\vspace{2mm}
	\noindent\textbf{Improvement in terms of training efficiency}
	Training ConvGNNs such as GCN \cite{kipf2017semi} usually is required to save the whole graph data and intermediate states of all nodes into memory. The full-batch training algorithm for ConvGNNs suffers significantly from the memory overflow problem, especially when a graph contains millions of nodes.  To save memory,  GraphSage \cite{hamilton2017inductive} proposes a batch-training algorithm for ConvGNNs. It samples a tree rooted at each node by recursively expanding the root node's neighborhood by $K$ steps with a fixed sample size. For each sampled tree, GraphSage computes the root node's hidden representation by hierarchically aggregating hidden node representations from bottom to top. %GraphSage has to sample an exponentially large neighborhood when $K$ becomes large. Suppose the number of neighbors to be sampled at each step is $r$, GraphSage needs to sample $r^K$ neighbors for each node. Therefore, GraphSage is suitable for the situation when $n$ is large and $K$ is small. 
	
	Fast Learning with Graph Convolutional Network (FastGCN) \cite{chen2018fastgcn} samples a fixed number of nodes for each graph convolutional layer instead of sampling a fixed number of neighbors for each node like GraphSage \cite{hamilton2017inductive}. It
    interprets graph convolutions as integral transforms of embedding functions of nodes under probability measures.  Monte Carlo approximation and variance reduction techniques are employed to facilitate the training process.  As FastGCN samples nodes independently for each layer,  between-layers connections are potentially sparse. Huang et al. \cite{huang2018ada} propose an adaptive layer-wise sampling approach where node sampling for the lower layer is conditioned on the top one. This method achieves higher accuracy compared to FastGCN at the cost of employing a much more complicated sampling scheme. 
    %This method is also applicable for explicit variance reduction. 
    
    In another work, Stochastic Training of Graph Convolutional Networks
    (StoGCN) \cite{chen2018stochastic} reduces the receptive field size of a graph convolution to an arbitrarily small scale using historical node representations as a control variate. StoGCN achieves comparable performance even with two neighbors per node. However, StoGCN still has to save intermediate states of all nodes, which is memory-consuming for large graphs.

    Cluster-GCN \cite{chiang2019cluster} samples a subgraph using a graph clustering algorithm and performs graph convolutions to nodes within the sampled subgraph. As the neighborhood search is also restricted within the sampled subgraph, Cluster-GCN is capable of handling larger graphs and using deeper architectures at the same time, in less time and with less memory.
    Cluster-GCN notably provides a straightforward comparison of time complexity and memory complexity for existing ConvGNN training algorithms. We analyze its results based on Table \ref{tab:complex}. 
    
    In Table \ref{tab:complex}, GCN \cite{kipf2017semi} is the baseline method which conducts the full-batch training. GraphSage saves memory at the cost of sacrificing time efficiency. Meanwhile, the time and memory complexity of GraphSage grows exponentially with an increase of $K$ and $r$. The time complexity of Sto-GCN is the highest, and the bottleneck of the memory remains unsolved. However, Sto-GCN can achieve satisfactory performance with very small $r$. The time complexity of Cluster-GCN remains the same as the baseline method since it does not introduce redundant computations. Of all the methods, Cluster-GCN realizes the lowest memory complexity.
    
    \vspace{2mm}
	\noindent\textbf{Comparison between spectral and spatial models}
    Spectral models have a theoretical foundation in graph signal processing. By designing new graph signal filters (e.g., Cayleynets \cite{levie2017cayleynets}), one can build new ConvGNNs. However, spatial models are preferred over spectral models due to efficiency, generality, and flexibility issues.
	First, spectral models are less efficient than spatial models. Spectral models either need to perform eigenvector computation or handle the whole graph at the same time.
	%, which makes them difficult to parallelize or scale to large graphs. 
	Spatial models are more scalable to large graphs as they directly perform convolutions in the graph domain via information propagation. The computation can be performed in a batch of nodes instead of the whole graph. 
	%When the number of neighboring nodes increases, sampling techniques \cite{hamilton2017inductive,chen2018fastgcn,huang2018ada,chiang2019cluster} can be developed to improve efficiency.
    Second, spectral models which rely on a graph Fourier basis generalize poorly to new graphs. They assume a fixed graph. Any perturbations to a graph would result in a change of eigenbasis.
    Spatial-based models, on the other hand, perform graph convolutions locally on each node where weights can be easily shared across different locations and structures.  
	Third, spectral-based models are limited to operate on undirected graphs. 
	%There is no clear definition of the Laplacian matrix on directed graphs so that the only way to apply spectral-based models to directed graphs is to transfer directed graphs to undirected graphs. 
	Spatial-based models are more flexible to handle multi-source graph inputs such as edge inputs \cite{scarselli2009graph,gilmer2017neural,kearnes2016molecular,pham2017column,simonovsky2017dynamic}, directed graphs \cite{atwood2016diffusion,li2018diffusion}, signed graphs \cite{derr2018signed}, and heterogeneous graphs \cite{such2017robust,wang2019heter}, because these graph inputs can be incorporated into the aggregation function easily.
	
    \subsection{Graph Pooling Modules}
    \label{sec:pooling}
    
    After a GNN generates node features, we can use them for the final task. But using all these features directly can be computationally challenging, thus, a down-sampling strategy is needed. Depending on the objective and the role it plays in the network, different names are given to this strategy: (1) the pooling operation aims to reduce the size of parameters by down-sampling the nodes to generate smaller representations and thus avoid overfitting, permutation invariance, and computational complexity issues; (2) the readout operation is mainly used to generate graph-level representation based on node representations. Their mechanism is very similar. In this chapter, we use pooling to refer to all kinds of down-sampling strategies applied to GNNs.
    
    In some earlier works, the graph coarsening algorithms use eigen-decomposition to coarsen graphs based on their topological structure. However, these methods suffer from the time complexity issue. The Graclus algorithm \cite{dhillon2007weighted} is an alternative of eigen-decomposition to calculate a clustering version of the original graph. Some recent works \cite{levie2017cayleynets} employed it as a pooling operation to coarsen graphs. 
    
    Nowadays, mean/max/sum pooling is the most primitive and effective way to implement down-sampling since calculating the mean/max/sum value in the pooling window is fast:
    \begin{equation}
    \label{eq:basicpool}
    \mathbf{h}_G = mean/max/sum(\mathbf{h}_1^{(K)}, \mathbf{h}_2^{(K)}, ..., \mathbf{h}_n^{(K)}),
    \end{equation}
    where $K$ is the index of the last graph convolutional layer.

    Henaff et al. \cite{henaff2015deep} show that performing a simple max/mean pooling at the beginning of the network is especially important to reduce the dimensionality in the graph domain and mitigate the cost of the expensive graph Fourier transform operation. Furthermore, some works \cite{li2015gated, gilmer2017neural, tran2018filter} also use attention mechanisms to enhance the mean/sum pooling.
    
    Even with attention mechanisms, the reduction operation (such as sum pooling) is not satisfactory since it makes the embedding inefficient: a fixed-size embedding is generated regardless of the graph size. Vinyals et al. \cite{vinyals2015order} propose the Set2Set method to generate a memory that increases with the size of the input. It then implements an LSTM that intends to integrate order-dependent information into the memory embedding before a reduction is applied that would otherwise destroy that information.
    
    % original 20/11/2019
    %Even with attention mechanisms, the reduction operation is inefficient and overlooks the impact of node ordering. Vinyals et al. \cite{vinyals2015order} propose the Set2Set method to reduce the order impact. It first embeds the input data into a memory vector, then feeds this vector into an LSTM for an update. After several iterations, a final embedding which is permutation invariant to the inputs will be generated. 
    
    %Gilmer et al. \cite{gilmer2017neural} employ the set2set mechanism as a pooling strategy to generate the graph level embedding which is invariant to the order of the node.} 
    
    Defferrard et al. \cite{defferrard2016convolutional} address this issue in another way by rearranging nodes of a graph in a meaningful way. They devise an efficient pooling strategy in their approach ChebNet. Input graphs are first coarsened into multiple levels by the Graclus algorithm \cite{dhillon2007weighted}. After coarsening, the nodes of the input graph and its coarsened version are rearranged into a balanced binary tree. Arbitrarily aggregating the balanced binary tree from bottom to top will arrange similar nodes together. Pooling such a rearranged signal is much more efficient than pooling the original.
    
    %Each node has either two children, if it was matched at the finer level, otherwise a fake node will be created to balance the tree.
    % original 29/11/2019
    %Defferrard et al. \cite{defferrard2016convolutional} address this issue in another way by optimizing the max/min pooling. They devise an efficient pooling strategy in their approach ChebNet. Input graphs are first processed by the Graclus algorithm. After coarsening, the nodes of the input graph and its coarsened versions are reformed in a balanced binary tree. Arbitrarily ordering the nodes at the coarsest level and propagating this ordering to a lower level in the balanced binary tree finally produces a regular ordering in the finest level. Pooling such a rearranged signal is much more efficient than pooling the original.
    
    Zhang et al. \cite{zhang2018end} propose the DGCNN with a similar pooling strategy named SortPooling which performs pooling by rearranging nodes to a meaningful order. Different from ChebNet \cite{defferrard2016convolutional}, DGCNN sorts nodes according to their structural roles within the graph. The graph's unordered node features from spatial graph convolutions are treated as continuous WL colors \cite{weisfeiler1968reduction}, and they are then used to sort nodes. In addition to sorting the node features, it unifies the graph size to $q$ by truncating/extending the node feature matrix. The last $n-q$ rows are deleted if $n>q$, otherwise $q-n$ zero rows are added. %This method is invariant to node permutations and thus enhances the performance of ConvGNNs.%This method enhances the pooling network to improve the performance of ConvGNNs by solving one challenge underlying graph-structured tasks which are referred to as permutation invariance.

    The aforementioned pooling methods mainly consider graph features and ignore the structural information of graphs. Recently, a differentiable pooling (DiffPool) \cite{ying2018hierarchical} is proposed, which can generate hierarchical representations of graphs. 
    %\textcolor{red}{which are important for capturing the structural information of a graph.} 
    Compared to all previous coarsening methods, DiffPool does not simply cluster the nodes in a graph but learns a cluster assignment matrix $\mathbf{S}$ at layer $k$ referred to as $\mathbf{S}^{(k)} \in \mathbf{R}^{n_k \times n_{k+1}}$, where $n_k$ is the number of nodes at the $k^{th}$ layer. The probability values in matrix $\mathbf{S}^{(k)}$ are being generated based on node features and topological structure using
	\begin{equation}
	\label{eq:difpool3}
	\mathbf{S}^{(k)} = softmax(ConvGNN_k(\mathbf{A}^{(k)}, \mathbf{H}^{(k)})).
	\end{equation}
    The core idea of this is to learn comprehensive node assignments which consider both topological and feature information of a graph, so Equation \ref{eq:difpool3} can be implemented with any standard ConvGNNs. %However, the drawback of this method is as obvious as it's great performance. While considering more information, the complexity of calculating the assignment matrix becomes a fatal limitation. 
    However, the drawback of DiffPool is that it generates dense graphs after pooling and thereafter the computational complexity becomes $O(n^2)$. 
    
    Most recently, the SAGPool \cite{lee2019self} approach is proposed, which considers both node features and graph topology and learns the pooling in a self-attention manner. 
    %Another approach, Graph U-net \cite{gao2019graph}, generalizes traditional architecture in computer vision to perform down-sampling in graphs. 
    
    Overall, pooling is an essential operation to reduce graph size. How to improve the effectiveness and computational complexity of pooling is an open question for investigation.
    %Overall, there are much few graph pooling methods than other graph neural networks with great improving space.
%    Overall, existing pooling methods either lack the topological and node feature information of graphs (mean/max/sum pooling, sort pooling, and Graclus pooling) or suffered from computation complexity (DiffPool). How to balance information integrity and computation complexity during pooling operation is still an open question for investigation.
	
	\subsection{Discussion of Theoretical Aspects}
	We discuss the theoretical foundation of graph neural networks from different perspectives.
	
	\vspace{2mm}
	\noindent\textbf{Shape of receptive field}
    The receptive field of a node is the set of nodes that contribute to the determination of its final node representation. When compositing multiple spatial graph convolutional layers, the receptive field of a node grows one step ahead towards its distant neighbors each time. Micheli  \cite{micheli2009neural} prove that a finite number of spatial graph convolutional layers exists such that for each node $v \in V$ the receptive field of node $v$ covers all nodes in the graph. As a result, a ConvGNN is able to extract global information by stacking local graph convolutional layers.  
    
	\vspace{2mm}
	\noindent\textbf{VC dimension}
	The VC dimension is a measure of model complexity defined as the largest number of points that can be shattered by a model. There are few works on analyzing the VC dimension of GNNs. Given the number of model parameter $p$ and the number of nodes $n$, Scarselli et al.  \cite{scarselli2018vapnik} derive that the VC dimension of a GNN* \cite{scarselli2009graph} is $O(p^4n^2)$ if it uses the sigmoid or tangent hyperbolic activation and is $O(p^2n)$ if it uses the piecewise polynomial activation function. This result
	suggests that the model complexity of a GNN* \cite{scarselli2009graph} increases rapidly with $p$ and $n$ if the sigmoid or tangent hyperbolic activation is used.

	\vspace{2mm}
	\noindent\textbf{Graph isomorphism} Two graphs are isomorphic if they are topologically identical. Given two non-isomorphic graphs $G_1$ and $G_2$, Xu et al. \cite{xu2019how} prove that if a GNN maps $G_1$ and $G_2$ to different embeddings, these two graphs can be identified as non-isomorphic by the Weisfeiler-Lehman (WL) test of isomorphism \cite{weisfeiler1968reduction}. They show that common GNNs such as GCN \cite{kipf2017semi} and GraphSage \cite{hamilton2017inductive} are incapable of distinguishing different graph structures. Xu et al. \cite{xu2019how} further prove if the aggregation functions and the readout functions of a GNN are injective, the GNN is at most as powerful as the WL test in distinguishing different graphs.

	\vspace{2mm}
	\noindent\textbf{Equivariance and invariance}
	A GNN must be an equivariant function when performing node-level tasks and must be an invariant function when performing graph-level tasks. For node-level tasks, let $f(\mathbf{A},\mathbf{X}) \in R^{n\times d}$ be a GNN and $\mathbf{Q}$ be any permutation matrix that changes the order of nodes. A GNN is equivariant if it satisfies $f(\mathbf{Q}\mathbf{A}\mathbf{Q}^T,\mathbf{Q}\mathbf{X})=\mathbf{Q}f(\mathbf{A},\mathbf{X})$. For graph-level tasks, let $f(\mathbf{A},\mathbf{X}) \in R^{d}$. A GNN is invariant if it satisfies $f(\mathbf{Q}\mathbf{A}\mathbf{Q}^T,\mathbf{Q}\mathbf{X})=f(\mathbf{A},\mathbf{X})$. In order to achieve equivariance or invariance, components of a GNN must be invariant to node orderings. Maron et al. \cite{maron2019invariant} theoretically study the characteristics of permutation invariant and equivariant linear layers for graph data.
	
	\vspace{2mm}
	\noindent\textbf{Universal approximation} It is well known that multi-perceptron feedforward neural networks with one hidden layer can approximate any Borel measurable functions \cite{hornik1989multilayer}. The universal approximation capability of GNNs has seldom been studied. Hammer et al. \cite{hammer2005universal} prove that cascade correlation can approximate functions with structured outputs. Scarselli et al. \cite{scarselli2008computational} prove that a RecGNN \cite{scarselli2009graph} can approximate any function that preserves unfolding equivalence up to any degree of precision. 	Two nodes are unfolding equivalent if their unfolding trees are identical where the unfolding tree of a node is constructed by iteratively extending a node's neighborhood at a certain depth. Xu et al. \cite{xu2019how} show that ConvGNNs under the framework of message passing \cite{gilmer2017neural} are not universal approximators of continuous functions defined on multisets. Maron et al. \cite{maron2019invariant} prove that an invariant graph network can approximate an arbitrary invariant function defined on graphs.

\begin{table*}[]
		\caption{Main characteristics of selected  GAEs}
		\label{tab:summary_gae}
		\centering
\begin{tabular}{lllll}
\toprule
Approaches & Inputs & Encoder & Decoder & Objective  \\ \toprule
  DNGR (2016) \cite{cao2016deep}         &  $A$      &    a multi-layer perceptron     &  a multi-layer perceptron       &  reconstruct the PPMI matrix           \\ \midrule
  SDNE (2016) \cite{wang2016sdne}         &  $A$      &   a multi-layer perceptron     &  a multi-layer perceptron      &        preserve node 1st-order and 2nd-order proximity       \\ \midrule
  GAE* (2016) \cite{kipf2016variational}          &  $A,X$      & a ConvGNN        & a similarity measure         &  reconstruct the adjacency matrix              \\ \midrule
  VGAE (2016) \cite{kipf2016variational}          &  $A,X$      &      a ConvGNN       & a similarity measure        &         learn the generative distribution of data       \\ \midrule
  ARVGA (2018) \cite{pan2018adversarially} &   $A,X$     &   a ConvGNN        & a similarity measure         &    learn the generative distribution of data adversarially            \\ \midrule
  DNRE (2018) \cite{tu2018deep} &   $A$     &  an LSTM network       &  an identity function       &  recover network embedding               \\ \midrule
  NetRA (2018) \cite{yu2018learning}&   $A$     &   an LSTM network       &    an LSTM network     &      recover network embedding with adversarial training          \\ \midrule
  DeepGMG (2018) \cite{li2018learning}&   $A,X,X^e$     &   a RecGNN      &  a decision process       &      maximize the expected joint log-likelihood           \\ \midrule
  GraphRNN (2018) \cite{you2018graphrnn}&    $A$    &  a RNN      &  a decision process       &        maximize the likelihood of permutations         \\ \midrule
  GraphVAE (2018) \cite{simonovsky2018graphvae}&  $A,X,X^e$     &  a ConvGNN       & a multi-layer perceptron       &    optimize the reconstruction loss            \\ \midrule
  RGVAE (2018) \cite{ma2018constrained}&   $A,X,X^e$      &   a CNN     & a deconvolutional net       &  optimize the reconstruction loss  with validity constraints              \\ \midrule
  MolGAN (2018) \cite{de2018molgan}&   $A,X,X^e$     &  a ConvGNN       &  a multi-layer perceptron      &     optimize the generative adversarial loss and the RL loss           \\ \midrule
  NetGAN (2018) \cite{bojchevski2018netgan}&   $A$     &    an LSTM network     &  an LSTM network       &          optimize the generative adversarial loss      \\ \midrule
\end{tabular}
\end{table*}

	\section{Graph autoencoders}
	\label{sec:genden}
	Graph autoencoders (GAEs) are deep neural architectures which map nodes into a latent feature space and decode graph information from  latent representations. GAEs can be used to learn network embeddings or generate new graphs. The main characteristics of selected GAEs are summarized in Table \ref{tab:summary_gae}. In the following, we provide a brief review of GAEs from two perspectives, network embedding and graph generation. 
	
	\subsection{Network Embedding}
	A network embedding is a low-dimensional vector representation of a node which preserves a node's topological information.  GAEs learn network embeddings using an encoder to extract network embeddings and using a decoder to enforce network embeddings to preserve the graph topological information such as the PPMI matrix and the adjacency matrix.
	
    Earlier approaches mainly employ multi-layer perceptrons to build GAEs for network embedding learning. Deep Neural Network for Graph Representations (DNGR) \cite{cao2016deep} uses a stacked denoising autoencoder \cite{vincent2008extracting} to encode and decode the PPMI matrix via multi-layer perceptrons. 
    %The learned latent representations are able to preserve highly non-linear regularity behind data and are robust especially when there are missing values. 
    %The PPMI matrix intrinsically captures nodes co-occurrence information through random walks sampled from a graph. 
	Concurrently, Structural Deep Network Embedding (SDNE) \cite{wang2016sdne} uses a stacked autoencoder to preserve the node first-order proximity and second-order proximity jointly. SDNE proposes two loss functions on the outputs of the encoder and the outputs of the decoder separately. The first loss function enables the learned network embeddings to preserve the node first-order proximity by minimizing the distance between a node's network embedding and its neighbors' network embeddings. The  first loss function $L_{1st}$ is defined as 
	\begin{equation}
	L_{1st} = \sum_{(v,u)\in E} A_{v,u}||enc(\mathbf{x}_v)-enc(\mathbf{x}_u)||^2,
	\end{equation}
	where $\mathbf{x}_v=\mathbf{A}_{v,:}$ and $enc(\cdot)$ is an encoder which consists of a multi-layer perceptron. The second loss function enables the learned network embeddings to preserve the node second-order proximity by minimizing the distance between a node's inputs and its reconstructed inputs. Concretely, the second loss function $L_{2nd}$ is defined as 
	\begin{equation}
	L_{2nd} = \sum_{v \in V} ||(dec(enc(\mathbf{x}_v))-\mathbf{x}_v)\odot \mathbf{b}_v||^2,
	\end{equation}
	where  $b_{v,u}=1$ if $A_{v,u}=0$, $b_{v,u}=\beta>1$ if $A_{v,u}=1$, and $dec(\cdot)$ is a decoder which consists of a multi-layer perceptron. 
	%The role of vector $b_i$ is to penalize non-zero elements more than zero elements since the inputs are highly sparse. 
	
    DNGR \cite{cao2016deep} and SDNE \cite{wang2016sdne} only consider node structural information which is about the connectivity between pairs of nodes. They ignore nodes may contain feature information that depicts the attributes of nodes themselves.  Graph Autoencoder (GAE*\footnote{We name it GAE* to avoid ambiguity in the survey.}) \cite{kipf2016variational} leverages GCN \cite{kipf2017semi} to encode node structural information and node feature information at the same time.  The encoder of GAE* consists of two graph convolutional layers, which takes the form
    \begin{equation}
        \label{eq:gae}
        \mathbf{Z} = enc(\mathbf{X},\mathbf{A}) = Gconv(f(Gconv(\mathbf{A},\mathbf{X};\mathbf{\Theta_1}));\mathbf{\Theta_2)},
    \end{equation}
	where $\mathbf{Z}$ denotes the network embedding matrix of a graph, $f(\cdot)$ is a ReLU activation function and the $Gconv(\cdot)$ function is a graph convolutional layer defined by Equation \ref{eq:1stchebnetc}. The decoder of GAE* aims to decode node relational information from their embeddings by reconstructing the graph adjacency matrix, which is defined as 
	\begin{equation}
	\hat{\mathbf{A}}_{v,u} = dec(\mathbf{z}_v,\mathbf{z}_u) = \sigma(\mathbf{z}_v^T\mathbf{z}_u),
	\end{equation}
	where $\mathbf{z}_v$ is the embedding of node $v$.
	GAE* is trained by minimizing the negative cross entropy given the real adjacency matrix $\mathbf{A}$ and the reconstructed adjacency matrix $\hat{\mathbf{A}}$.
	
    Simply reconstructing the graph adjacency matrix may lead to overfitting due to the capacity of the autoencoders.  Variational Graph Autoencoder (VGAE) \cite{kipf2016variational} is a variational version of GAE to learn the distribution of data. VGAE optimizes the variational lower bound $L$:
	
	\begin{equation}
	    \label{eq:vgae}
	    L = E_{q(\mathbf{Z}|\mathbf{X},\mathbf{A})}[\log p(\mathbf{A}|\mathbf{Z})]-KL[q(\mathbf{Z}|\mathbf{X},\mathbf{A})||p(\mathbf{Z})],
	\end{equation}
	where $KL(\cdot)$ is the Kullback-Leibler divergence function which measures the distance between two distributions, $p(\mathbf{Z})$ is a Gaussian prior $p(\mathbf{Z})=\prod_{i=1}^np(\mathbf{z}_i)=\prod_{i=1}^nN(\mathbf{z}_i|0,\mathbf{I})$,  $p(A_{ij}=1|\mathbf{z}_i,\mathbf{z}_j)=dec(\mathbf{z}_i,\mathbf{z}_j)=\sigma(\mathbf{z}_i^T\mathbf{z}_j)$, $q(\mathbf{Z}|\mathbf{X},\mathbf{A})=\prod_{i=1}^nq(\mathbf{z}_i|\mathbf{X},\mathbf{A})$  with $q(\mathbf{z}_i|\mathbf{X},\mathbf{A})=N(\mathbf{z}_i|\mathbf{\mu}_i,diag(\mathbf{\sigma}_i^2))$. The mean vector $\mathbf{\mu}_i$ is the $i^{th}$ row of an encoder's outputs defined by Equation \ref{eq:gae} and $\log \mathbf{\sigma}_i$ is derived similarly as $\mathbf{\mu}_i$ with another encoder. According to Equation \ref{eq:vgae}, VGAE assumes the empirical distribution $q(\mathbf{Z}|\mathbf{X},\mathbf{A})$ should be as close as possible to the prior distribution $p(\mathbf{Z})$.
    To further enforce the empirical distribution $q(\mathbf{Z}|\mathbf{X},\mathbf{A})$ approximate the prior distribution $p(\mathbf{Z})$, Adversarially Regularized Variational Graph Autoencoder (ARVGA)
	\cite{pan2018adversarially,pan2019learning} employs the training scheme of a generative adversarial networks (GAN) \cite{goodfellow2014generative}. A GAN plays a competition game between a generator and a discriminator in training generative models. A generator tries to generate `fake samples' to be as real as possible while a discriminator attempts to distinguish the `fake samples' from real ones. Inspired by GANs, ARVGA endeavors to learn an encoder that produces an empirical distribution $q(\mathbf{Z}|\mathbf{X},\mathbf{A})$ which is indistinguishable from the prior distribution $p(\mathbf{Z})$.

	Similar as GAE*, GraphSage \cite{hamilton2017inductive} encodes node features with two graph convolutional layers. 
	Instead of optimizing the reconstruction error, GraphSage shows that the relational information between two nodes can be preserved by negative sampling with the loss:
	\begin{equation}
	    L(\mathbf{z}_v) = -log(dec(\mathbf{z}_v,\mathbf{z}_u))-QE_{v_n\sim P_n(v)}\log (-dec(\mathbf{z}_v,\mathbf{z}_{v_n})),
	\end{equation}
	where node $u$ is a neighbor of node $v$, node $v_n$ is a distant node to node $v$ and is sampled from a negative sampling distribution $P_n(v)$, and Q is the number of negative samples. This loss function essentially enforces close nodes to have similar representations and distant nodes to have dissimilar representations. DGI \cite{velivckovic2019deep} alternatively drives local network embeddings to capture global structural information by maximizing local mutual information. It shows a distinct improvement over GraphSage \cite{hamilton2017inductive} experimentally.

    For the aforementioned methods, they essentially learn network embeddings by solving a link prediction problem. However, the sparsity of a graph causes the number of positive node pairs to be far less than the number of negative node pairs. To alleviate the data sparsity problem in learning network embedding, another line of works convert a graph into sequences by random permutations or random walks. In such a way, those deep learning approaches which are applicable to sequences can be directly used to process graphs.
	Deep Recursive Network Embedding (DRNE) \cite{tu2018deep} assumes a node's network embedding should approximate the aggregation of its neighborhood network embeddings. It adopts a Long Short-Term Memory (LSTM) network \cite{hochreiter1997long} to aggregate a node's neighbors. The reconstruction error of DRNE is defined as 
	\begin{equation}
	\label{eq:drne}
	    L = \sum_{v\in V} ||\mathbf{z}_v-LSTM(\{\mathbf{z}_u|u\in N(v)\})||^2,
	\end{equation}
	where $\mathbf{z}_v$ is the network embedding of node $v$ obtained by a dictionary look-up, and the LSTM network takes a random sequence of node $v$'s neighbors ordered by their node degree as inputs. As suggested by Equation \ref{eq:drne}, DRNE implicitly learns network embeddings via an LSTM network rather than using the LSTM network to generate network embeddings. It avoids the problem that the LSTM network is not invariant to the permutation of node sequences. 
	Network Representations with Adversarially Regularized Autoencoders (NetRA) \cite{yu2018learning} proposes a graph encoder-decoder framework with a general loss function, defined as
    \begin{equation}
        L= -E_{\mathbf{z}\sim P_{data}(\mathbf{z})}(dist(\mathbf{z},dec(enc(\mathbf{z})))),
    \end{equation}
    where $dist(\cdot)$ is the distance measure between the node embedding $\mathbf{z}$ and the reconstructed $\mathbf{z}$. The encoder and decoder of NetRA are LSTM networks with random walks rooted on each node $v\in V$ as inputs.
    Similar to ARVGA \cite{pan2018adversarially}, NetRA regularizes the learned network embeddings within a prior distribution via adversarial training. Although NetRA ignores the node permutation variant problem of LSTM networks, the experimental results validate the effectiveness of NetRA. 
	%Since NetRA is not invariant to permutations of nodes, the learned representation of each node is not deterministic.  
	
	\subsection{Graph Generation}
	With multiple graphs, GAEs are able to learn the generative distribution of graphs by encoding graphs into hidden representations and decoding a graph structure given hidden representations. The majority of GAEs for graph generation are designed to solve the molecular graph generation problem, which has a high practical value in drug discovery. These methods either propose a new graph in a sequential manner or in a global manner.  
	
	Sequential approaches generate a graph by proposing nodes and edges step by step. Gomez et al. \cite{gomez2018automatic}, Kusner et al. \cite{kusner2017grammar}, and Dai et al. \cite{dai2018syntax} model the generation process of a string representation of molecular graphs named SMILES with deep CNNs and RNNs as the encoder and the decoder respectively. While these methods are domain-specific, alternative solutions are applicable to general graphs by means of iteratively adding nodes and edges to a growing graph until a certain criterion is satisfied.  
	Deep Generative Model of Graphs (DeepGMG) \cite{li2018learning}
	assumes the probability of a graph is the sum over all possible node permutations:
	\begin{equation}
	    p(G)=\sum_\pi p(G,\pi),
	\end{equation}
	where $\pi$ denotes a node ordering. It captures the complex joint probability of all nodes and edges in the graph.
	DeepGMG generates graphs by making a sequence of decisions, namely whether to add a node, which node to add, whether to add an edge, and which node to connect to the new node. The decision process of generating nodes and edges is conditioned on the node states and the graph state of a growing graph updated by a RecGNN.  In another work, GraphRNN \cite{you2018graphrnn} proposes a graph-level RNN and an edge-level RNN to model the generation process of nodes and edges.
    The graph-level RNN adds a new node to a node sequence each time while the edge-level RNN produces a binary sequence indicating connections between the new node and the nodes previously generated in the sequence. 
    
    %CGVAE (Constrained Graph Variational Autoencoders) \cite{liu2018constrained} assumes the generative probability of a graph is the sum over all possible node permutations. Since it is intractable, CGVAE proposes a variational autoencoder with GGNNs \cite{li2015gated} 
    
    Global approaches output a graph all at once.  Graph Variational Autoencoder (GraphVAE) \cite{simonovsky2018graphvae} models the existence of nodes and edges as independent random variables. By assuming the posterior distribution $q_\mathbf{\phi}(\mathbf{z}|G)$ defined by an encoder and the generative distribution $p_\theta(G|\mathbf{z})$ defined by a decoder, GraphVAE optimizes the variational lower bound:
    \begin{equation}
L(\mathbf{\phi},\mathbf{\theta};G)= E_{q_\mathbf{\phi}(z|G)}[-\log p_\theta(G|\mathbf{z})]+KL[q_\mathbf{\phi}(\mathbf{z}|G)||p(\mathbf{z})],
    \end{equation}
    where $p(\mathbf{z})$ follows a Gaussian prior, $\mathbf{\phi}$ and $\mathbf{\theta}$ are learnable parameters. With a ConvGNN as the encoder and a simple multi-layer perception as the decoder, GraphVAE outputs a generated graph with its adjacency matrix, node attributes and edge attributes. It is challenging to control the global properties of generated graphs, such as graph connectivity, validity, and node compatibility.  Regularized Graph Variational Autoencoder (RGVAE) \cite{ma2018constrained} further imposes validity constraints on a graph variational autoencoder to regularize the output distribution of the decoder. 
    Molecular Generative Adversarial Network (MolGAN) \cite{de2018molgan} integrates convGNNs \cite{schlichtkrull2018modeling}, GANs \cite{gulrajani2017improved} and reinforcement learning objectives to generate graphs with the desired properties. MolGAN consists of a generator and a discriminator, competing with each other to improve the authenticity of the generator. In MolGAN, the generator tries to propose a fake graph along with its feature matrix while the discriminator aims to distinguish the fake sample from the empirical data. Additionally, a reward network is introduced in parallel with the discriminator to encourage the generated graphs to possess certain properties according to an external evaluator. 
    NetGAN \cite{bojchevski2018netgan} combines LSTMs \cite{hochreiter1997long} with Wasserstein GANs \cite{arjovsky2017wasserstein} to generate graphs from a random-walk-based approach. NetGAN trains a generator to produce plausible random walks through an LSTM network and enforces a discriminator to identify fake random walks from the real ones. After training, a new graph is derived by normalizing a co-occurrence matrix of nodes computed based on random walks produced by the generator. 
    
    In brief, sequential approaches linearize graphs into sequences. They can lose structural information due to the presence of cycles.  Global approaches produce a graph all at once. They are not scalable to large graphs as the output space of a GAE is up to $O(n^2)$.

\section{Spatial-temporal Graph Neural Networks}
\label{sec:stgcn}
Graphs in many real-world applications are dynamic both in terms of graph structures and graph inputs. Spatial-temporal graph neural networks (STGNNs) occupy important positions in capturing the dynamicity of graphs. Methods under this category aim to model the dynamic node inputs while assuming interdependency between connected nodes. For example, a traffic network consists of speed sensors placed on roads where edge weights are determined by the distance between pairs of sensors. As the traffic condition of one road may depend on its adjacent roads' conditions, it is necessary to consider spatial dependency when performing traffic speed forecasting.
As a solution,  STGNNs capture spatial and temporal dependencies of a graph simultaneously. The task of STGNNs can be forecasting future node values or labels, or predicting spatial-temporal graph labels.  STGNNs follow two directions, RNN-based methods and CNN-based methods.
	
Most RNN-based approaches capture spatial-temporal dependencies by filtering inputs and hidden states passed to a recurrent unit using graph convolutions \cite{zhang2018gaan, seo2018structured, li2018diffusion}. To illustrate this, suppose a simple RNN takes the form
\begin{equation}
    \label{eq:rnn}
    \mathbf{H}^{(t)} = \sigma(\mathbf{W}\mathbf{X}^{(t)}+\mathbf{U}\mathbf{H}^{(t-1)}+\mathbf{b}), 
\end{equation}
where $\mathbf{X}^{(t)}\in \mathbf{R}^{n\times d}$ is the node feature matrix at time step $t$. After inserting graph convolution, Equation \ref{eq:rnn} becomes
\begin{equation}
    \label{eq:rnn1}
    \mathbf{H}^{(t)} = \sigma(Gconv(\mathbf{X}^{(t)},\mathbf{A};\mathbf{W})+Gconv(\mathbf{H}^{(t-1)},\mathbf{A};\mathbf{U})+\mathbf{b}),
\end{equation}
where $Gconv(\cdot)$ is a graph convolutional layer. 
Graph Convolutional Recurrent Network (GCRN) \cite{seo2018structured} combines a LSTM network with ChebNet \cite{defferrard2016convolutional}. Diffusion Convolutional Recurrent Neural Network (DCRNN) \cite{li2018diffusion} incorporates a proposed diffusion graph convolutional layer (Equation \ref{eq:dcn}) into a GRU network. In addition, DCRNN adopts an encoder-decoder framework to predict the future $K$ steps of node values. 

Another parallel work uses node-level RNNs and edge-level RNNs to handle different aspects of temporal information. Structural-RNN \cite{jain2016structural} proposes a recurrent framework to predict node labels at each time step. It comprises two kinds of RNNs, namely a node-RNN and an edge-RNN. The temporal information of each node and each edge is passed through a node-RNN and an edge-RNN respectively. To incorporate the spatial information, a node-RNN takes the outputs of edge-RNNs as inputs. Since assuming different RNNs for different nodes and edges significantly increases model complexity, it instead splits nodes and edges into semantic groups. 
%For example, a human-object interaction graph consists of two groups of nodes, human nodes and object nodes, and three groups of edges, human-human edges, object-object edges, and human-object edges. 
Nodes or edges in the same semantic group share the same RNN model, which saves the computational cost. %However, Structural-RNN demands human prior knowledge to split the semantic groups. 
	
	%\subsubsection{GCN Based  spatial-temporal graph networks}
	%To meet the demands of multi-step forecasting, DCRNN adopts sequence-to-sequence architecture \cite{sutskever2014sequence} where the recurrent unit is replaced by DCGRU.
	
RNN-based approaches suffer from time-consuming iterative propagation and gradient explosion/vanishing issues. As alternative solutions, CNN-based approaches tackle spatial-temporal graphs in a non-recursive manner with the advantages of parallel computing, stable gradients, and low memory requirements. As illustrated in Fig \ref{fig:gst}, CNN-based approaches interleave 1D-CNN layers with graph convolutional layers to learn temporal and spatial dependencies respectively. Assume the inputs to a spatial-temporal graph neural network is a tensor $\mathbf{\mathcal{X}}\in R^{T\times n\times d}$, the 1D-CNN layer slides over $\mathbf{\mathcal{X}}_{[:,i,:]}$ along the time axis to aggregate temporal information for each node while the graph convolutional layer operates on $\mathbf{\mathcal{X}}_{[i,:,:]}$ to aggregate spatial information at each time step.  CGCN \cite{yuspatio} integrates 1D convolutional layers with ChebNet \cite{defferrard2016convolutional} or GCN \cite{kipf2017semi} layers. It constructs a spatial-temporal block by stacking a gated 1D convolutional layer, a graph convolutional layer and another gated 1D convolutional layer in a sequential order. ST-GCN \cite{yan2018spatial} composes a spatial-temporal block using a 1D convolutional layer and a PGC layer (Equation \ref{eq:pgcn}). 

Previous methods all use a pre-defined graph structure. They assume the pre-defined graph structure reflects the genuine dependency relationships among nodes. However,  with many snapshots of graph data in a spatial-temporal setting, it is possible to learn latent static graph structures automatically from data. To realize this, Graph WaveNet \cite{wu2019graph} proposes a self-adaptive adjacency matrix to perform graph convolutions. The self-adaptive adjacency matrix is defined as 
\begin{equation}
    \mathbf{A}_{adp}=SoftMax(ReLU(\mathbf{E}_1\mathbf{E}_2^T)),
\end{equation}
where the SoftMax function is computed along the row dimension, $\mathbf{E1}$ denotes the source node embedding and $\mathbf{E2}$ denotes the target node embedding with learnable parameters. By multiplying $\mathbf{E1}$ with $\mathbf{E2}$, one can get the dependency weight between a source node and a target node.
With a complex CNN-based spatial-temporal neural network,  Graph WaveNet performs well without being given an adjacency matrix.

Learning latent static spatial dependencies can help researchers discover interpretable and stable correlations among different entities in a network. However, in some circumstances, learning latent dynamic spatial dependencies may further improve model precision. For example, in a traffic network, the travel time between two roads may depend on their current traffic conditions. GaAN \cite{zhang2018gaan} employs attention mechanisms to learn dynamic spatial dependencies through an RNN-based approach. An attention function is used to update the edge weight between two connected nodes given their current node inputs.  
ASTGCN \cite{guo2019att} further includes a spatial attention function and a temporal attention function to learn latent dynamic spatial dependencies and temporal dependencies through a CNN-based approach. 
The common drawback of learning latent spatial dependencies is that it needs to calculate the spatial dependency weight between each pair of nodes, which costs $O(n^2)$.

%In order to capture long-term temporal dependencies, CGCN and ST-GCN have to stack many spatial-temporal blocks together or use a global pooling layer in the end. 

	%\subsubsection{Miscellaneous Variants}
	%\vspace{2mm}

	\section{Applications}\label{sec:applications}
	As graph-structured data are ubiquitous, GNNs have a wide variety of applications. In this section, we  summarize the benchmark graph data sets, evaluation methods, and open-source implementation, respectively. We detail practical applications of GNNs in various domains.

	\subsection{Data Sets}
    %Data sets are indispensable in evaluating the performance of GNNs across different tasks such as node classification and graph classification.  
    We mainly sort data sets into four groups, namely citation networks, biochemical graphs, social networks, and others.  In Table \ref{tab:datafreq}, we summarize selected benchmark data sets. More details is given in the Supplementary Material A.

	\begin{table*}[htbp]
		\caption{Summary of selected benchmark data sets.}
		\label{tab:datafreq}
		\centering
		% \usepackage{multirow}
		% \usepackage{multirow}
		\begin{tabular}{l l l l l l l l l }
			\toprule
			\multicolumn{1}{l}{Category} & Data set & Source &\# Graphs & \# Nodes(Avg.) & \#     Edges (Avg.) & \#Features & \# Classes & Citation \\ 
			\midrule
			\hline
			\multicolumn{1}{l|}{\multirow{4}{*}{\begin{tabular}[c]{@{}l@{}}Citation\\Networks \end{tabular}}} & Cora &\cite{sen2008collective} &1 & 2708 & 5429 & 1433 & 7 &  \begin{tabular}[c]{@{}l@{}}
			   \cite{kipf2017semi,levie2017cayleynets,atwood2016diffusion,zhuang2018dual,velickovic2017graph,monti2017geometric,gao2018large}\\ 
			   \cite{chen2018fastgcn,chen2018stochastic,huang2018ada,li2018deeper,velivckovic2019deep,kipf2016variational,pan2018adversarially}\\ 
			\end{tabular}\\ \cline{2-9} 
			\multicolumn{1}{l|}{} & Citeseer & \cite{sen2008collective} &1 & 3327 & 4732 & 3703 & 6 &        \begin{tabular}[c]{@{}l@{}}\cite{kipf2017semi,zhuang2018dual,velickovic2017graph,gao2018large,chen2018stochastic,huang2018ada,li2018deeper}\\
			\cite{velivckovic2019deep,kipf2016variational,pan2018adversarially}\\

			\end{tabular}                                                  \\ \cline{2-9} 
			\multicolumn{1}{l|}{}                                   & Pubmed                          &          \cite{sen2008collective}                                                                  &1       & 19717           & 44338    & 500        & 3         &            \begin{tabular}[c]{@{}l@{}} \cite{dai2018learning,kipf2017semi,atwood2016diffusion,zhuang2018dual,velickovic2017graph,monti2017geometric,gao2018large}\\
			\cite{chen2018fastgcn,huang2018ada,li2018deeper,liu2018geniepath,velivckovic2019deep,kipf2016variational,pan2018adversarially}\\ \cite{bojchevski2018netgan,pham2017column}\\
			\end{tabular}
			\\ \cline{2-9} 
			\multicolumn{1}{l|}{}                                   & DBLP (v11)                            &          \begin{tabular}[c]{@{}l@{}}\cite{tang2008arnetminer}\\\end{tabular}         &1                                                                & 4107340           & 36624464    &        -    & -         &   \cite{yu2018learning,bojchevski2018netgan,wang2019heter}                                                      \\ \hline

			%\multicolumn{1}{l|}{}                                   & COLLAB                            &                 \cite{yanardag2015deep}              &                          5000                          &        74.49         &   4914.99       &       -     &    3       &                 \begin{tabular}[c]{@{}l@{}}\cite{tran2018filter,zhang2018end,xu2019how} \\\end{tabular}                                     \\ \cline{2-9}
			%\multicolumn{1}{l|}{}                                   & IMDB-B                             &                 \cite{yanardag2015deep}              &                               1000                     &        284.31         &   715.65       &       -     &    2       &                 \begin{tabular}[c]{@{}l@{}}\cite{tran2018filter,zhang2018end,xu2019how} \\\end{tabular}                                     \\ \cline{2-9}
			%\multicolumn{1}{l|}{}                                   & IMDB-M                            &                 \cite{yanardag2015deep}              &                 1500                                   &        13         &   131.87       &       -     &    3       &                 \begin{tabular}[c]{@{}l@{}}\cite{tran2018filter,zhang2018end,xu2019how} \\\end{tabular}                                     \\ \hline

			\multicolumn{1}{l|}{\multirow{8}{*}{\begin{tabular}[c]{@{}l@{}}Bio-\\chemical \\Graphs\end{tabular}}}   & PPI                             &  \cite{zitnik2017predicting}     & 24                                                                            & 56944 & 818716        & 50         & 121       &              \begin{tabular}[c]{@{}l@{}}\cite{dai2018learning,hamilton2017inductive,velickovic2017graph,zhang2018gaan,gao2018large,chen2018stochastic,liu2018geniepath}\\\cite{velivckovic2019deep,chiang2019cluster,yu2018learning}\end{tabular}                                           \\ \cline{2-9} 
			\multicolumn{1}{l|}{}                                   & NCI-1                           &              \cite{wale2008comparison}      &           4110                                                    &    29.87             &     32.30     &   37        &    2       &       \begin{tabular}[c]{@{}l@{}}\cite{atwood2016diffusion,niepert2016learning,tran2018filter,zhang2018end,xu2019how,simonovsky2017dynamic,such2017robust} \end{tabular}                                               \\ \cline{2-9} 
			%\multicolumn{1}{l|}{}                                   & NCI-109                         &       \cite{wale2008comparison}         &       4127                                                            &    29.6             &     -     &      38      &      2     &        \cite{atwood2016diffusion,niepert2016learning,simonovsky2017dynamic}                                                 \\ \cline{2-9} 
			\multicolumn{1}{l|}{}                                   & MUTAG                         &   \cite{debnath1991structure}             &          188                                                         &        17.93         &   19.79       &       7     &      2     &   \begin{tabular}[c]{@{}l@{}}\cite{atwood2016diffusion,niepert2016learning,tran2018filter,zhang2018end,xu2019how,simonovsky2017dynamic}    \\ \end{tabular}                                                \\ \cline{2-9} 
			%to do: proteins \cite{xu2019how,verma2018graph,ying2018hierarchical,zhang2018end,tran2018filter}
			%to do: ptc \cite{xu2019how,verma2018graph,zhang2018end,tran2018filter}
			%to do: ENZYMES \cite{ying2018hierarchical,verma2018graph}
			\multicolumn{1}{l|}{}                                   & D\&D                             &                 \cite{dobson2003distinguishing}              &          1178                                          &        284.31         &   715.65      &       82     &    2       &                 \begin{tabular}[c]{@{}l@{}}\cite{niepert2016learning,tran2018filter,zhang2018end,ying2018hierarchical,simonovsky2017dynamic,such2017robust}  \end{tabular}                                     \\ \cline{2-9}
			
			\multicolumn{1}{l|}{}                                   & PROTEIN                             &                 \cite{borgwardt2005protein}              &        1113                                            &        39.06         &   72.81       &       4     &    2       &                 \begin{tabular}[c]{@{}l@{}}\cite{niepert2016learning,tran2018filter,zhang2018end,ying2018hierarchical,xu2019how} \\ \end{tabular}                                     \\ \cline{2-9}
			
			\multicolumn{1}{l|}{}                                   & PTC                             &                 \cite{toivonen2003statistical}              &          344                                          &        25.5         &   -       &       19     &    2       &                 \begin{tabular}[c]{@{}l@{}}\cite{atwood2016diffusion,niepert2016learning,tran2018filter,zhang2018end,xu2019how} \\ \end{tabular}   
			
			\\ \cline{2-9}

			\multicolumn{1}{l|}{}                                   & QM9       &       \cite{ramakrishnan2014quantum}               &                133885                                                                   &          -       &      -    &       -     &     -      &      \cite{gilmer2017neural,de2018molgan}                                                   \\
			\cline{2-9}
			
			\multicolumn{1}{l|}{}                                   & Alchemy                             &                 \cite{chen2019alchemy}              &                      119487                              &       -        &   -       &       -     &    -       &  -
			\\ \hline
			%\multicolumn{1}{l|}{}                                   & tox21    &   \begin{tabular}[c]{@{}l@{}}tripod.nih.gov/\\tox21/challenge/\end{tabular}                    &       12707                                                                            &         -        &    -      &     -       &    12       &         \cite{li2018adaptive,kearnes2016molecular}                                                \\  \hline
			
			%\multicolumn{1}{l|}{\multirow{3}{*}{\begin{tabular}[c]{@{}l@{}}Unstruct-\\ured \\Graphs\end{tabular}}} & MNIST                   &  \begin{tabular}[c]{@{}l@{}}yann.lecun.com\\/exdb/mnist/\end{tabular}       &  70000                                                                      &       -          &     -     &       -     &     10      &             \cite{levie2017cayleynets,defferrard2016convolutional,bruna2013spectral,simonovsky2017dynamic,monti2017geometric}                                            \\ \cline{2-9} 
			%\multicolumn{1}{l|}{} & Wikipedia                       &  \begin{tabular}[c]{@{}l@{}}www.mattmahoney\\.net/dc/textdata\end{tabular}& 1 & 4777            & 184812   & -          & 40        & \cite{yu2018learning} \\ \cline{2-9} 
			%\multicolumn{1}{l|}{} & 20NEWS                    &   \cite{joachims1996probabilistic}      &   1                                                                       &      18846           &    -      &   -         &    20       &          \cite{defferrard2016convolutional,cao2016deep}                                               \\ \hline
			\multicolumn{1}{l|}{\multirow{2}{*}{\begin{tabular}[c]{@{}l@{}}Social \\Networks\end{tabular}}}     & Reddit                          &                     \cite{hamilton2017inductive}       &1                                                       &    232965             &     11606919     &   602         &    41       &     \begin{tabular}[c]{@{}l@{}}\cite{hamilton2017inductive,zhang2018gaan,chen2018fastgcn,chen2018stochastic,huang2018ada,velivckovic2019deep}  \\  \end{tabular}                                                 \\ \cline{2-9}

            \multicolumn{1}{l|}{}   & BlogCatalog                     &               \cite{tang2009relational}                             & 1                                       & 10312           & 333983   &   -         & 39        &     \cite{dai2018learning,liu2018geniepath,wang2016sdne,yu2018learning}                                                    \\ \hline
			\multicolumn{1}{l|}{\multirow{3}{*}{Others}} & MNIST                   &  \cite{lecun1998gradient}     &  70000                                                                    &       784          &     -     &       1     &     10      &             \cite{bruna2013spectral,levie2017cayleynets,defferrard2016convolutional,monti2017geometric,simonovsky2017dynamic}                                            \\ \cline{2-9} 
			\multicolumn{1}{l|}{} & METR-LA                    &      \cite{jagadish2014big}  &       1                                                                    &        207         &     1515     &     2       &   -        &   \cite{zhang2018gaan,li2018diffusion,wu2019graph}                                                      \\ \cline{2-9} 
			%\multicolumn{1}{l|}{} & Movie-Lens1M                    &  \begin{tabular}[c]{@{}l@{}}\cite{miller2003movielens}\\grouplens.org/\\datasets/\\movielens/1m/\end{tabular}     &       1                                                                     &      10000           &      1 Million    &     -       &      -     &   \cite{levie2017cayleynets}                                                      \\ \cline{2-9} 
			\multicolumn{1}{l|}{}                                   & Nell                            &   \cite{carlson2010toward}     &        1                                                                   & 65755           & 266144   & 61278      & 210       &    \cite{kipf2017semi,zhuang2018dual,chen2018stochastic}                                                   \\
			\hline
			\bottomrule
		\end{tabular}
	\end{table*}

    \subsection{Evaluation \& Open-source Implementations}
    Node classification and graph classification are common tasks to assess the performance of RecGNNs and ConvGNNs. %Most existing methods evaluate the performance either on a fix split of  train/valid/test for node classification or 10 fold cross validation for graph classification. However, these evaluation may not necessarily represent a rigorous comparison. 
    
    \vspace{1mm}
    \textbf{Node Classification} In node classification, most methods follow a standard split of train/valid/test on benchmark data sets including Cora, Citeseer, Pubmed, PPI, and Reddit.  They reported the average accuracy or F1 score on the test data set over multiple runs. A summarization of experimental results of methods can be found in the Supplementary Material B.  It should be noted that these results do not necessarily represent a rigorous comparison. 
    Shchur et al. identified \cite{shchur2018pitfalls} two pitfalls in evaluating the performance GNNs on node classification. First, using the same train/valid/test split throughout all experiments underestimates the generalization error. Second, different methods employed different training techniques such as hyper-parameter tuning, parameter initialization, learning rate decay, and early stopping. For a relatively fair comparison, we refer the readers to  Shchur et al. \cite{shchur2018pitfalls}. 
    
    \vspace{1mm}
    \textbf{Graph Classification} In graph classification, researchers often adopt 10-fold cross validation (cv) for model evaluation. However, as pointed out by \cite{anonymous2020a}, the experimental settings are ambiguous and not unified across different works. In particular, \cite{anonymous2020a} raises the concern of the correct usage of data splits for model selection versus model assessment. An often encountered problem is that the external test set of each fold is used both for model selection and risk assessment. \cite{anonymous2020a} compare GNNs in a standardized and uniform evaluation framework. They apply an external 10 fold CV to get an estimate of the generalization performance of a model and  an inner holdout technique with a 90\%/10\% training/validation split for model selection. An alternative procedure would be a double cv method, which uses an external $k$ fold cv for model assessment and an inner $k$ fold cv for model selection. We refer the readers to \cite{anonymous2020a} for a detailed and rigorous comparison of GNN methods for graph classification.
%Due to the number of methods being compared is limited, we do not provide experimental results for graph classification in our survey.
    
%    Table \ref{tab:bennode} collects the experimental results of methods which followed this evaluation process. It should noted that Table \ref{tab:bennode} does not represent a rigorous comparison. Shchur et al. identified \cite{shchur2018pitfalls} two pitfalls in evaluating the performance GNNs on node classification. First, using the same train/valid/test split throughout all experiments underestimates the generalization error. Second, different methods employed different training techniques such as hyper-parameter tuning, parameter initialization, learning rate decay, and early stopping. For a relatively fair comparison, we refer readers to  Shchur et al.'s work \cite{shchur2018pitfalls}. In graph classification, researchers often adopt 10-fold cross validation for model evaluation. However, as pointed out by \cite{}, the experimental settings are ambiguous and not unified across different works. In particular, \cite{} raises the concern of the correct usage of data splits for model selection vesus model assessment. A often encountered problem is that the external test set of each fold is used both for model selection and risk assessment. \cite{} compare GNNs in a standardized and uniform evaluation framework for graph classification. They adopt a double cross validation procedure. Due to the number of methods being compared is limited, we do not provide experimental results for graph classification in our survey.

    \vspace{1mm}
    \textbf{Open-source implementations} facilitate the work of baseline experiments in deep learning research.  In the Supplementary Material C, we provide the hyperlinks of the open-source implementations of the GNN models reviewed in this paper. Noticeably, Fey et al. \cite{fey2018splinecnn} published a geometric learning library in PyTorch named PyTorch Geometric \footnote{\url{https://github.com/rusty1s/pytorch_geometric}}, which implements many GNNs. Most recently, the Deep Graph Library (DGL) \footnote{https://www.dgl.ai/} \cite{wang2019dgl} is released which provides a fast implementation of many GNNs on top of popular deep learning platforms such as PyTorch and MXNet. 

    \subsection{Practical Applications}
    GNNs have many applications across different tasks and domains. Despite general tasks which can be handled by each category of GNNs directly, including node classification, graph classification, network embedding, graph generation, and spatial-temporal graph forecasting, other general graph-related tasks such as node clustering \cite{wang2017mgae}, link prediction \cite{zhang2018link}, and graph partitioning \cite{kawamoto2018mean} can also be addressed by GNNs.  We detail some applications based on the following research domains.%As most of these general tasks have been discussed previously, we mainly introduce practical applications based on research domains in this section. 
    %These applications, on the other hand, fall into one of the following tasks. In this section, we give a brief summary of each domain.
    
    \vspace{2mm}\noindent\textbf{Computer vision}
    Applications of GNNs in computer vision include scene graph generation, point clouds classification, and action recognition.
    
    Recognizing semantic relationships between objects facilitates the understanding of the meaning behind a visual scene. Scene graph generation models aim to parse an image into a semantic graph which consists of objects and their semantic relationships \cite{xu2017scene,yang2018graph,li2018factorizable}.   Another application inverses the process by generating realistic images given scene graphs \cite{johnson2018image}. As natural language can be parsed as semantic graphs where each word represents an object, it is a promising solution to synthesize images given textual descriptions. 
    
    Classifying and segmenting points clouds enables LiDAR devices to `see' the surrounding environment. A point cloud is a set of 3D points recorded by LiDAR scans.  \cite{wang2018dynamic,landrieu2017large, te2018rgcnn} convert point clouds into k-nearest neighbor graphs or superpoint graphs and use ConvGNNs to explore the topological structure.
    
    Identifying human actions contained in videos facilitates a better understanding of video content from a machine aspect. Some solutions detect the locations of human joints in video clips. Human joints which are linked by skeletons naturally form a graph. Given the time series of human joint locations, \cite{jain2016structural,yan2018spatial} apply STGNNs to learn human action patterns.
    
    Moreover, the number of applicable directions of GNNs in computer vision is still growing. It includes human-object interaction \cite{qi2018learning}, few-shot image classification \cite{garcia2018fewshot, guo2018neural,liu2019prototype}, semantic segmentation \cite{qi20173d,yi2017syncspeccnn}, visual reasoning \cite{chen2018iterative}, and question answering \cite{narasimhan2018out}.

    \vspace{2mm}\noindent\textbf{Natural language processing} 
    %is an interdisciplinary between computers and human languages. It endeavors to process, analyze, and understand natural language data from a machine perspective. 
    A common application of GNNs in natural language processing is text classification. GNNs utilize the inter-relations of documents or words to infer document labels \cite{kipf2017semi, hamilton2017inductive, velickovic2017graph}.

    Despite the fact that natural language data exhibit a sequential order,  they may also contain an internal graph structure, such as a syntactic dependency tree. A syntactic dependency tree defines the syntactic relations among words in a sentence. Marcheggiani et al. \cite{marcheggiani2017encoding} propose the Syntactic GCN which runs on top of a CNN/RNN sentence encoder. The Syntactic GCN aggregates hidden word representations based on the syntactic dependency tree of a sentence. Bastings et al.  \cite{bastings2017graph} apply the Syntactic GCN to the task of neural machine translation. Marcheggiani et al. \cite{marcheggiani2018exploiting} further adopt the same model as Bastings et al.  \cite{bastings2017graph} to handle the semantic dependency graph of a sentence. %Experiments show that incorporating syntactic relations performs better than incorporating semantic relations in neural machine translation. However, considering both syntactic relations and semantic relations leads to higher performance.

    Graph-to-sequence learning learns to generate sentences with the same meaning given a semantic graph of abstract words (known as Abstract Meaning Representation). Song et al. \cite{song2018graph} propose a graph-LSTM to encode graph-level semantic information. Beck et al.  \cite{beck2018graph} apply a GGNN \cite{li2015gated} to graph-to-sequence learning and neural machine translation.
    The inverse task is sequence-to-graph learning. Generating a semantic or knowledge graph given a sentence is very useful in knowledge discovery \cite{johnson2016learning,chen2018sequence}.

    \vspace{2mm}\noindent\textbf{Traffic} Accurately forecasting traffic speed, volume or the density of roads in traffic networks is fundamentally important in a smart transportation system.   \cite{zhang2018gaan,li2018diffusion,yuspatio} address the traffic prediction problem using STGNNs. They consider the traffic network as a spatial-temporal graph where the nodes are sensors installed on roads, the edges are measured by the distance between pairs of nodes, and each node has the average traffic speed within a window as dynamic input features.  Another industrial-level application is taxi-demand prediction. 
    %This greatly helps intelligent transportation systems make use of resources and save energy effectively.  
    Given historical taxi demands, location information, weather data, and event features,  Yao et al. \cite{yao2018deep} incorporate LSTM, CNN and network embeddings trained by LINE \cite{tang2015line} to form a joint representation for each location to predict the number of taxis demanded for a location within a time interval.

    \vspace{2mm}\noindent\textbf{Recommender systems}
    Graph-based recommender systems take items and users as nodes. By leveraging the relations between items and items, users and users, users and items, as well as content information, graph-based recommender systems are able to produce high-quality recommendations. The key to a recommender system is to score the importance of an item to a user. As a result, it can be cast as a link prediction problem. To predict the missing links between users and items, Van et al. \cite{van2017graph} and Ying et al. \cite{ying2018graph} propose a GAE which uses ConvGNNs as encoders. Monti et al. \cite{monti2017geometricb} combine RNNs with graph convolutions to learn the underlying process that generates the known ratings.

    \vspace{2mm}\noindent\textbf{Chemistry} In the field of chemistry, researchers apply GNNs to study the graph structure of molecules/compounds. In a molecule/compound graph, atoms are considered as nodes, and chemical bonds are treated as edges. Node classification, graph classification,  and graph generation are the three main tasks targeting molecular/compound graphs in order to  learn molecular fingerprints \cite{duvenaud2015convolutional,kearnes2016molecular}, to predict molecular properties \cite{gilmer2017neural}, to infer protein interfaces \cite{fout2017protein}, and to synthesize chemical compounds \cite{li2018learning,de2018molgan,you2018graph}.
	
	%\textbf{Molecular classification and generation}

	\vspace{2mm}\noindent\textbf{Others} The application of GNNs is not limited to the aforementioned domains and tasks.
	There have been explorations of applying GNNs to a variety of problems such as program verification \cite{li2015gated}, program reasoning \cite{allamanis2017learning}, social influence prediction \cite{qiu2018deepinf}, adversarial attacks prevention \cite{zugner2018adversarial}, electrical health records modeling \cite{choi2017gram,choi2018mime}, brain networks \cite{kawahara2017brainnetcnn}, event detection \cite{nguyen2018graph}, and combinatorial optimization \cite{li2018combinatorial}.

	\section{Future Directions}\label{sec:fucture}
	Though GNNs have proven their power in learning graph data,  challenges still exist due to the complexity of graphs. In this section, we suggest four future directions of GNNs.
	
	\vspace{2mm}\noindent\textbf{Model depth}
	The success of deep learning lies in deep neural architectures \cite{he2016deep}.   However, Li et al. show that the performance of a ConvGNN drops dramatically with an increase in the number of graph convolutional layers \cite{li2018deeper}. As graph convolutions push representations of adjacent nodes closer to each other, in theory, with an infinite number of graph convolutional layers, all nodes' representations will converge to a single point \cite{li2018deeper}.
	This raises the question of whether going deep is still a good strategy for learning graph data.
	
%	\vspace{2mm}\noindent\textbf{Receptive Field} The receptive field of a node refers to a set of nodes including the central node and its neighbors. The number of neighbors of a node follows a power law distribution. Some nodes may only have one neighbor, while other nodes may have as many as thousands of neighbors.  Though sampling strategies have been adopted \cite{hamilton2017inductive,niepert2016learning,gao2018large}, how to select a representative receptive field of a node remains to be explored. 
	
	\vspace{2mm}\noindent\textbf{Scalability trade-off} The scalability of GNNs is gained at the price of corrupting graph completeness.  Whether using sampling or clustering, a model will lose part of the graph information.  By sampling, a node may miss its influential neighbors. By clustering,  a graph may be deprived of a distinct structural pattern. How to trade-off algorithm scalability and graph integrity could be a future research direction.
	
	\vspace{2mm}\noindent\textbf{Heterogenity}
	The majority of current GNNs assume homogeneous graphs.  It is difficult to directly apply current GNNs to heterogeneous graphs, which may contain different types of nodes and edges, or different forms of node and edge inputs,  such as images and text. Therefore, new methods should be developed to handle heterogeneous graphs.
	
	\vspace{2mm}\noindent\textbf{Dynamicity}
	Graphs are in nature dynamic in a way that nodes or edges may appear or disappear, and that node/edge inputs may change time by time. New graph convolutions are needed to adapt to the dynamicity of graphs. Although the dynamicity of graphs can be partly addressed by STGNNs,  few of them consider how to perform graph convolutions in the case of dynamic spatial relations.

	%The majority of current GNNs tackle with static homogeneous graphs.On the one hand, graph structures are assumed to be fixed. On the other hand, nodes and edges from a graph are assumed to come from a single source.  However, these two assumptions are not realistic in many scenarios. In a social network, a new person may enter into a network at any time, and an existing person may quit the network as well. In a recommender system, products may have different types where their inputs may have different forms such as texts or images. Therefore, new methods should be developed  to handle dynamic and heterogeneous graph structures.

	\section{Conclusion}\label{sec:conclusion}
	In this survey, we conduct a comprehensive overview of graph neural networks. We provide a taxonomy which groups graph neural networks into four categories: recurrent graph neural networks, convolutional graph neural networks, graph autoencoders, and spatial-temporal graph neural networks. We provide a thorough review, comparisons, and summarizations of the methods within or between categories.   Then we introduce a wide range of applications of graph neural networks. Data sets, open-source codes, and model assessment for graph neural networks are summarized. Finally, we suggest four future directions for graph neural networks.

 	\section*{Acknowledgment}	
 	This research was funded by the Australian Government through the Australian Research Council (ARC) under grants 1) LP160100630 partnership with Australia Government Department of Health and 2) LP150100671 partnership with Australia Research Alliance for Children and Youth (ARACY) and Global Business College Australia (GBCA). %We  acknowledge the support of NVIDIA Corporation and MakeMagic Australia with the donation of GPU used for this research.

\bibliographystyle{IEEEtran}
\bibliography{IEEEabrv,biblio}

%\balance

%\input{bio.tex}
%\pagebreak
\section*{Appendix}
%\section{Supplemental Material}

\subsection{Data Set}
	\vspace{2mm}\noindent\textbf{Citation Networks}
	consist of papers, authors, and their relationships such as citations, authorship, and co-authorship.
	Although citation networks are directed graphs, they are often treated as undirected graphs in evaluating model performance with respect to node classification, link prediction, and node clustering tasks.  There are three popular data sets for paper-citation networks, Cora, Citeseer and Pubmed.  The Cora data set contains 2708 machine learning publications grouped into seven classes.  The Citeseer data set contains 3327 scientific papers grouped into six classes. Each paper in Cora and Citeseer is represented by a one-hot vector indicating the presence or absence of a word from a dictionary.  The Pubmed data set contains 19717 diabetes-related publications. Each paper in Pubmed is represented by a term frequency-inverse document frequency (TF-IDF) vector. Furthermore, DBLP is a large citation data set with millions of papers and authors which are collected from computer science bibliographies. The raw data set of DBLP can be found on \url{https://dblp.uni-trier.de}.  A processed version of the DBLP paper-citation network is updated continuously by \url{https://aminer.org/citation}. 
	
		\begin{table}[h]
	\caption{Reported experimental results for node classification on five frequently used data sets.  Cora, Citeseer, and Pubmed are evaluated by classification accuracy.
		PPI and Reddit are evaluated by micro-averaged F1 score. 
		%For fair comparison, the listed methods use standard splits of these data sets. %In detail, for Cora, the standard split is [train,valid,test]=[140,500,1000] (nodes). For Citeseer, the standard split is [train,valid,test]=[120,500,1000] (nodes). For Pubmed, the standard split is [train,valid,test]=[600,500,1000] (nodes). For PPI, [train,valid,test]=[20,2,2] (graphs). Cora, 
	}
	\label{tab:bennode}
	\centering
	\scriptsize
	\begin{tabular}{l l l l l l}
		\toprule
		Method       & Cora     & Citeseer & Pubmed   & PPI   & Reddit    \\ \midrule
		SSE (2018)          & -        & -        & -        & 83.60    & - \\ \hline
		GCN (2016)   & 81.50     & 70.30     & 79.00     & -     &  -  \\ \hline
		%SGC (2019) \cite{wu2019simplifying} & 81.00 & 71.90 & 78.90 & - & \\ \hline
		Cayleynets (2017)   & 81.90 & -        & -        & -   &    - \\ \hline
		DualGCN (2018)      & 83.50     & 72.60     & 80.00     & -    &  -   \\ \hline
		GraphSage (2017)     & -        & -        & -        & 61.20     &  95.40 \\ \hline
		GAT (2017)          & 83.00 & 72.50 & 79.00 &  97.30 & -  \\ \hline
		MoNet (2017)  & 81.69   & -    &  78.81    & -     &   -  \\ \hline
		LGCN (2018)         & 83.30 & 73.00 & 79.50 & 77.20  &  -  \\ \hline
		GAAN (2018)         & -        & -        & -        & 98.71 & 96.83 \\ \hline
		FastGCN (2018)  & - & - & - & - & 93.70 \\ \hline
		StoGCN (2018)     & 82.00 & 70.90  & 78.70   & 97.80   &  96.30 \\ \hline
		Huang et al. (2018) & - & - & -& - & 96.27 \\ \hline
		GeniePath (2019)   & -        & -        & 78.50     & 97.90      & - \\ \hline
		DGI (2018) & 82.30 & 71.80 & 76.80 & 63.80 & 94.00 \\ \hline
		Cluster-GCN (2019) & - & - &- & 99.36 & 96.60 \\ \bottomrule
	\end{tabular}
\end{table}

\begin{table*}
	\caption{A Summary of Open-source Implementations}
	\label{tab:codes}
	\centering
	\begin{tabular}{ l l l }
		\toprule
		Model & Framework & Github Link \\ \midrule
		GGNN (2015)      &    torch         &      \url{https://github.com/yujiali/ggnn}       \\ \hline
		SSE (2018)      &    c         &      \url{https://github.com/Hanjun-Dai/steady_state_embedding}       \\ \hline
		ChebNet (2016)   & tensorflow  &\url{https://github.com/mdeff/cnn_graph}    \\ \hline 
		GCN (2017) & tensorflow  &\url{https://github.com/tkipf/gcn}          \\ \hline
		
		CayleyNet (2017) & tensorflow & \url{https://github.com/amoliu/CayleyNet}.  \\ \hline 
		DualGCN (2018)  & theano & \url{https://github.com/ZhuangCY/DGCN} \\ \hline
		GraphSage (2017)     &  tensorflow           &        \url{https://github.com/williamleif/GraphSAGE}     \\ \hline
		GAT (2017)          &     tensorflow        &    \url{https://github.com/PetarV-/GAT}         \\ \hline 
		LGCN (2018)         &    tensorflow         &      \url{https://github.com/divelab/lgcn/}       \\ \hline
		PGC-DGCNN (2018) & pytorch & \url{https://github.com/dinhinfotech/PGC-DGCNN} \\ \hline
		%SGC (2019) \cite{wu2019simplifying} & pytorch & \url{https://github.com/Tiiiger/SGC} \\ \hline
		FastGCN (2018)  & tensorflow & \url{https://github.com/matenure/FastGCN} \\ \hline
		StoGCN (2018)  & tensorflow & \url{https://github.com/thu-ml/stochastic_gcn} \\ \hline
		DGCNN (2018)  & torch & \url{https://github.com/muhanzhang/DGCNN} \\ \hline
		DiffPool (2018)  & pytorch & \url{https://github.com/RexYing/diffpool} \\ \hline
		DGI (2019)  & pytorch & \url{https://github.com/PetarV-/DGI} \\ \hline
		GIN (2019)  & pytorch & \url{https://github.com/weihua916/powerful-gnns} \\ \hline
		Cluster-GCN (2019)  & pytorch & \url{https://github.com/benedekrozemberczki/ClusterGCN} \\ \hline
		%CapsGNN (2019) \cite{xinyi2019capsule} & pytorch & \url{https://github.com/benedekrozemberczki/CapsGNN} \\ \hline
		DNGR (2016)        &      matlab       & \url{https://github.com/ShelsonCao/DNGR}            \\ \hline 
		SDNE (2016)      & tensorflow            &    \url{https://github.com/suanrong/SDNE}         \\ \hline 
		GAE (2016)        &  tensorflow           &    \url{https://github.com/limaosen0/Variational-Graph-Auto-Encoders}         \\ \hline 
		ARVGA (2018)          &  tensorflow           &    \url{https://github.com/Ruiqi-Hu/ARGA}         \\ \hline 
		DRNE (2016)        &      tensorflow       & \url{https://github.com/tadpole/DRNE}            \\ \hline
		GraphRNN (2018)      &    tensorflow         &    \url{https://github.com/snap-stanford/GraphRNN}         \\ \hline  
		MolGAN (2018) & tensorflow & \url{https://github.com/nicola-decao/MolGAN} \\ \hline
		NetGAN (2018) & tensorflow & \url{https://github.com/danielzuegner/netgan}  \\ \hline
		GCRN (2016)  & tensorflow & \url{https://github.com/youngjoo-epfl/gconvRNN} \\ \hline
		DCRNN (2018)     &     tensorflow        &      \url{https://github.com/liyaguang/DCRNN}        \\ \hline 
		Structural RNN (2016) &     theano        & \url{https://github.com/asheshjain399/RNNexp}            \\ \hline
		CGCN (2017)        &   tensorflow          &    \url{https://github.com/VeritasYin/STGCN_IJCAI-18}        \\ \hline 
		ST-GCN (2018)        & pytorch           &   \url{https://github.com/yysijie/st-gcn}          \\
		\hline
		GraphWaveNet (2019) & pytorch & \url{https://github.com/nnzhan/Graph-WaveNet} \\ \hline
		
		ASTGCN (2019)  & mxnet & \url{https://github.com/Davidham3/ASTGCN} \\ \bottomrule
	\end{tabular}
\end{table*}

	\vspace{2mm}\noindent\textbf{Biochemical Graphs}
	Chemical molecules and compounds can be represented by chemical graphs with atoms as nodes and chemical bonds as edges.  This category of graphs is often used to evaluate graph classification performance.  The NCI-1 and NCI-9 data set contain 4110 and 4127 chemical compounds respectively, labeled as to whether they are active to hinder the growth of human cancer cell lines. The MUTAG data set contains 188 nitro compounds, labeled as to whether they are aromatic or heteroaromatic. The D\&D and PROTEIN data set represent proteins as graphs, labeled as to whether they are enzymes or non-enzymes.
	The PTC data set consists of 344 chemical compounds, labeled as to whether they are carcinogenic for male and female rats. 
	The QM9 data set records 13 physical properties of 133885 molecules with up to 9 heavy atoms. The Alchemy data set records 12 quantum mechanical properties of 119487 molecules comprising up to 14 heavy atoms. %The Tox21 data set contains 12707 chemical compounds labeled with 12 types of toxicity. 
	Another important data set is the Protein-Protein Interaction network (PPI). It contains 24 biological graphs with nodes represented by proteins and edges represented by the interactions between proteins. In PPI, each graph is associated with one human tissue. Each node is labeled with its biological states.
	
	%\vspace{2mm}\noindent\textbf{Unstructured Graphs}
	%To test the generalization of graph neural networks to unstructured data, the $k$ nearest neighbor graph (k-NN graph) has been widely used. The MNIST data set contains  70000 images of size $28\times 28$ labeled with ten digits. A typical way to convert an MNIST image to a graph is to construct an 8-NN graph based on its pixel locations. 
	%The Wikipedia data set is a word co-occurrence network extracted from the first million bytes of the Wikipedia dump. Labels of words represent part-of-speech (POS) tags.  
	%The 20-NewsGroup data set consists of around 20,000 News Group (NG) text documents categorized by 20 news types.   The graph of the 20-NewsGroup is constructed by representing each document as a node and using the similarities between nodes as edge weights.
	\vspace{2mm}\noindent\textbf{Social Networks}
	are formed by user interactions from online services such as BlogCatalog and Reddit. The BlogCatalog data set is a social network which consists of bloggers and their social relationships.  The classes of bloggers represent their personal interests.  The Reddit data set is an undirected graph formed by posts collected from the Reddit discussion forum. Two posts are linked if they contain comments by the same user. Each post has a label indicating the community to which it belongs. 
	
	\vspace{2mm}\noindent\textbf{Others}
	There are several other data sets worth mentioning. The MNIST data set contains  70000 images of size $28\times 28$ labeled with ten digits. An MNINST image is converted to a graph by constructing an 8-nearest-neighbors graph based on its pixel locations.
	The METR-LA is a spatial-temporal graph data set. It contains four months of traffic data collected by 207 sensors on the highways of Los Angeles County. The adjacency matrix of the graph is computed by the sensor network distance with a Gaussian threshold. 
	%The MovieLens-1M data set from the MovieLens website contains 1 million item ratings given by 6k users.  It is a benchmark data set for recommender systems. 
	The NELL data set is a knowledge graph obtained from the Never-Ending Language Learning project. It consists of facts represented by a triplet which involves two entities and their relation.

\subsection{Reported Experimental Results for Node Classification}
A summarization of experimental results of methods which follow a standard train/valid/test split is given in Table \ref{tab:bennode}.

\subsection{Open-source Implementations}
Here we summarize the open-source implementations of graph neural networks reviewed in the survey. We provide the hyperlinks of the source codes of the GNN models in table \ref{tab:codes}.

\balance
\end{document}